\documentclass[10pt, conference, letterpaper]{IEEEtran}
\IEEEoverridecommandlockouts
\usepackage[utf8]{inputenc}
\usepackage[style=ieee]{biblatex}
\usepackage{amsmath,amssymb,amsfonts}
\usepackage{stfloats}
\usepackage[ruled,vlined,linesnumbered]{algorithm2e}
\usepackage{graphicx}
\usepackage{textcomp}
\usepackage{xcolor}
\usepackage{multirow}
\usepackage{caption}
\usepackage{subcaption}
\usepackage{array}

\begin{document}
\title{Cost Effective MLaaS Federation: A Combinatorial Reinforcement Learning Approach}

\author{\IEEEauthorblockN{Shuzhao Xie\IEEEauthorrefmark{2}, Yuan Xue\IEEEauthorrefmark{2}, Yifei Zhu\IEEEauthorrefmark{3}, Zhi Wang\IEEEauthorrefmark{2}\IEEEauthorrefmark{4}\IEEEauthorrefmark{1}\thanks{\IEEEauthorrefmark{1}Corresponding author.}} 
\IEEEauthorblockA{\IEEEauthorrefmark{2}Shenzhen International Graduate School, Tsinghua University}
\IEEEauthorblockA{\IEEEauthorrefmark{3}UM-SJTU Joint Institute, Shanghai Jiao Tong University}
\IEEEauthorblockA{\IEEEauthorrefmark{4}Peng Cheng Laboratory}
\IEEEauthorblockA{\{jsz20, xuey21\}@mails.tsinghua.edu.cn, yifei.zhu@sjtu.edu.cn, wangzhi@sz.tsinghua.edu.cn}
}

\maketitle

\begin{abstract}
With the advancement of deep learning techniques, major cloud providers and niche machine learning service providers start to offer their cloud-based machine learning tools, also known as machine learning as a service (MLaaS), to the public. According to our measurement, for the same task, these MLaaSes from different providers have varying performance due to the proprietary datasets, models, etc. Federating different MLaaSes together allows us to improve the analytic performance further. However, naively aggregating results from different MLaaSes not only incurs significant momentary cost but also may lead to sub-optimal performance gain due to the introduction of possible false-positive results. In this paper, we propose Armol, a framework to federate the right selection of MLaaS providers to achieve the best possible analytic performance. We first design a word grouping algorithm to unify the output labels across different providers. We then present a deep combinatorial reinforcement learning based-approach to maximize the accuracy while minimizing the cost. The predictions from the selected providers are then aggregated together using carefully chosen ensemble strategies. The real-world trace-driven evaluation further demonstrates that Armol is able to achieve the same accuracy results with $67\%$ less inference cost.

\end{abstract}


\section{Introduction} \label{introduction}
Recent advancements in machine learning techniques and the maturation of cloud services have propelled the introduction of machine learning as a service, in which cloud providers offer machine learning training platforms or machine learning inference services via machine learning APIs to users. 
Major cloud providers, such as Amazon Web Service\footnote{https://aws.amazon.com} (AWS), Microsoft Azure\footnote{https://azure.microsoft.com}, Google Cloud Platform\footnote{https://cloud.google.com} (GCP), etc., and niche machine learning vendors, such as BigML\footnote{https//bigml.com}, Algorithmia\footnote{https://algorithmia.com/}, etc., have all offered their own MLaaS. The MLaaS market was valued at $1.60$ billion USD in $2020$ and is expected to reach $12.10$ billion USD by $2026$ \cite{mlaasmarket}. The well-defined interfaces and the free maintenance burden for the underlying cloud infrastructures allow more industrial verticals and applications to access the machine learning process from anywhere, at any time.


From the users’ perspective, although the high abstraction of MLaaS brings ease of use, these abstracted services have also made the underlying latency, accuracy unknown to the users. To explore the underlying mechanisms of cloud services, previous works mainly focus on measuring the inference accuracy and latency of user-known models \cite{reddi2020mlperf, zhang2020inferbench}. The performance of cloud-based inference services has not been studied yet. 
According to our initial measurement by collecting the predictions from object detection services, we ﬁnd that the overall mAP of the major cloud services signiﬁcantly differs from each other, and each provider has different sweet-spot categories of tasks that achieve the best analytic performance than other categories. For example, in our measurement, though AWS outperforms Azure by $3.7\%$ in the general object detection task, Azure outperforms AWS by $10.9\%$ in a specific ``bottle'' category. Therefore, leveraging the service provider with the highest general accuracy loses the opportunities to fully exploit the analytic capability of all providers. On the contrary, aggregating all service providers may also introduce extra false-positive results. To gain the most from these MLaaS providers, it is thus beneficial to combine the expertise from different service providers and select the right set of MLaaS providers.

However, realizing MLaaS federation is non-trivial. First, different MLaaS providers may have the different vocabulary to describe the same task. With the fast update of each provider's services, we need an efficient algorithm to unify the description languages used in different providers.
Second, it is computational challenging to select the right set of providers due to the combinatorial nature of this problem. The large possible provider list and the resulting exponential number of choices make the brute-force approach not scalable and practical in real-world scenarios. Third, after receiving the analytic results from different service providers, how to efficiently merge these results to offer optimal aggregated results also need further design.

Therefore, in this paper, we present Armol, the first work on MLaaS federation for optimal analytic performance. Our framework covers three parts: the provider selection part, the word grouping part, and the ensemble part. Specially, we propose a combinatorial reinforcement learning (RL)-based approach to solve the provider selection problem. We map continuous action spaces to discrete integer action spaces by finding the nearest neighbor in large discrete combinatorial action spaces of a continuous action so that we can solve the computational challenge of selecting the right providers.
Our word grouping part unifies the categories with the same meaning from different providers based on the synonym dataset extracted from WordNet \cite{10.1145/219717.219748}.
In the ensemble part, we use an affirmative voting strategy and weighted box fusion for ablation so that the total analytic results can be further optimized. 
We conduct extensive real trace-driven experiments to evaluate the performance of our framework. 


In summary, our contributions are: 
\begin{itemize}
\item Our measurement studies on major cloud providers reveal the varying differences among existing MLaaS offerings and the great potential in MLaaS federation to improve analytic performance.
\item We formulate the MLaaS federation problem as a combinatorial provider selection problem and propose a combinatorial reinforcement learning-based approach to maximize accuracy. 
\item Efficient ensemble and grouping strategies are proposed to unify the vocabulary of different providers and aggregate the eventual results. 
\item Real-world trace-driven simulations demonstrate that our framework can reduce $67\%$ cost of inference fee without sacrificing accuracy compared to other benchmark approaches.
\end{itemize}

The remainder of this paper is organized as follows. Sec.~\ref{measurement} explains why MLaaS federation is necessary and possible. Sec.~\ref{problemformulation} describes the MLaaS federation problem formulation and explains why we need a combinatorial RL approach. Sec.~\ref{systemdesign} introduces the three parts of Armol. We evaluate Armol in Sec.~\ref{evaluation}. Sec.~\ref{relatedwork} presents the related work, followed by the conclusion in Sec.~\ref{conclusion}.

\section{Measurement \& Motivation} \label{measurement}
In this section, we analyze the latency and accuracy of existing major MLaaS products, AWS Rekognition \cite{awsrekog}, Azure Computer Vision \cite{azurecv}, and Google Cloud Vision AI \cite{googlecv}, and demonstrate the great benefit in MLaaS federation and the feasibility for achieving this.

We conduct the measurement from March to July $2021$ and rent the virtual machines (VMs) located in Singapore and the USA from AWS and Azure as clients to request these major MLaaS products. The types of AWS VMs are \texttt{t1.micro} and \texttt{t2.micro}, and the type of Azure VMs is \texttt{Standard B2s}. The above VMs have similar CPU, memory, storage, and network bandwidth. We request these services via Python SDK and capture the TCP packets by \texttt{tcpdump}. We take the object detection task as an example, and the accuracy metrics selected for this task are mean average precision ($mAP$), $mAP$ with intersection over union (IoU) threshold $50\%$ ($AP_{50}$), and $mAP$ with IoU threshold $75\%$ ($AP_{75}$) \cite{mapvoc2012}. COCO Val $2017$ \cite{lin2014microsoft} is chosen to test the performance of these services. For AWS Rekognition and Azure Computer Vision, we choose Singapore as the region for cloud service because the users prefer to choose the closest region to reduce the latency. Unlike the above two, Google Cloud Vision AI picks the region for the user, who cannot choose the region themselves. 

\begin{table}[t]
\caption{$AP$ of different MLaaS providers.}
\begin{center}
\begin{tabular}{|c|c|c|c|}
\hline
 \textbf{Provider} & $\mathbf{mAP}$ & $\mathbf{AP_{50}}$ & $\mathbf{AP_{75}}$ \\ 
\hline
 AWS & $18.81$ & $28.88$ & $20.84$ \\  
 Azure & $15.10$ & $24.38$ & $16.14$ \\
 GCP & $16.23$ & $23.03$ & $18.12$ \\
\hline
\end{tabular}
\label{mapofdiffods}
\end{center}
\end{table}

\begin{figure}[!t]
\centerline{\includegraphics[width=0.48\textwidth]{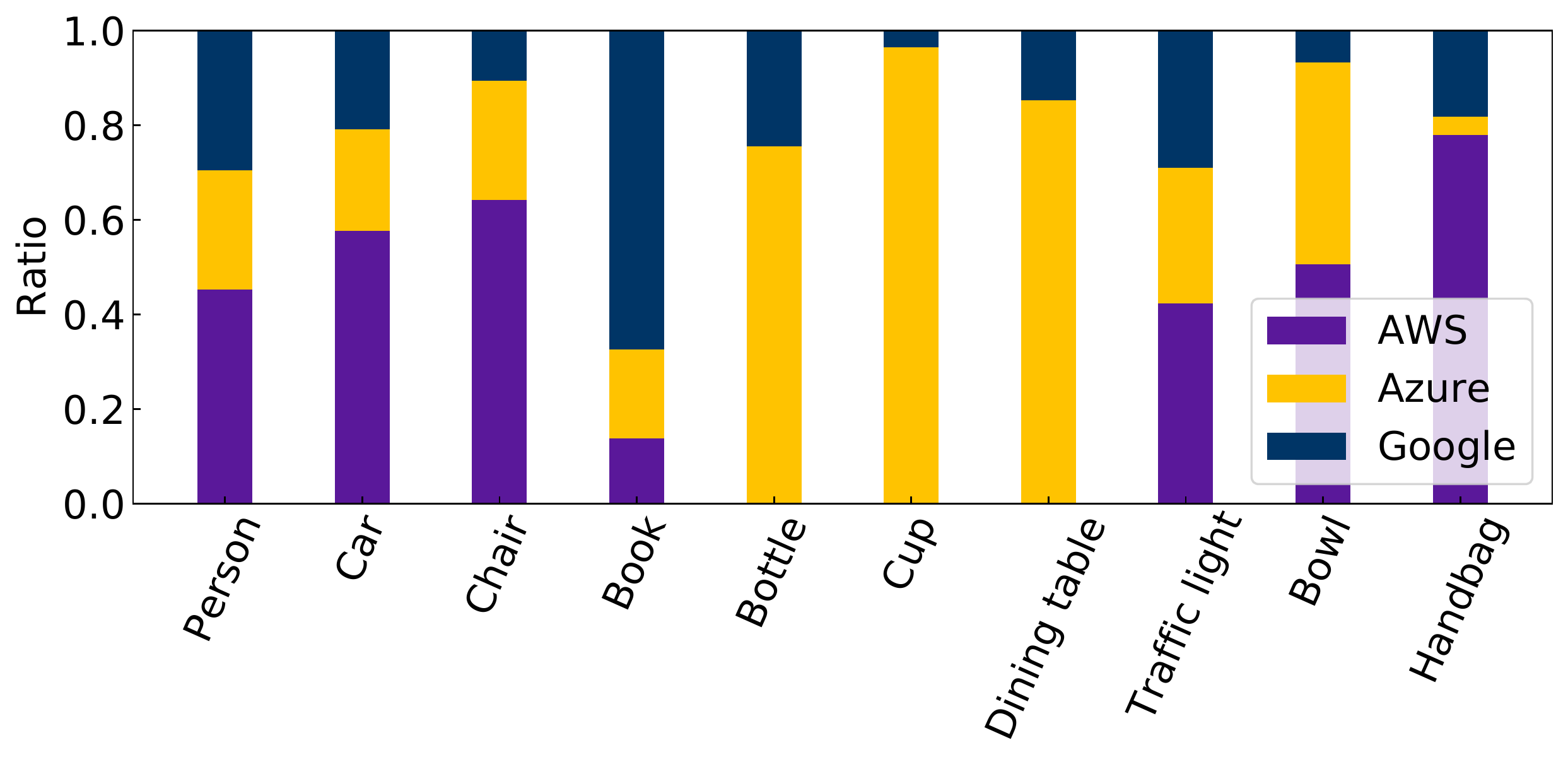}}
\caption{The AP comparison of AWS Rekognition, Azure Computer Vision, and Google Cloud Vision AI on top-$10$ frequent categories of COCO Val $2017$.}
\label{fig:top_10_cmp}
\end{figure}

\begin{figure*}[!t]
\centering
\begin{subfigure}[b]{0.24\textwidth}
\centerline{\includegraphics[width=\textwidth]{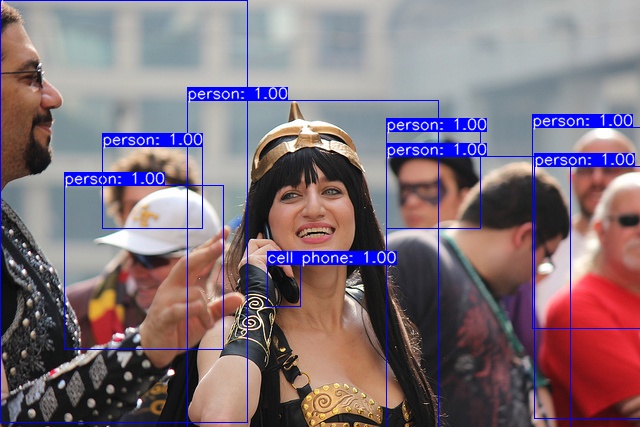}}
\caption{Ground truth, $AP_{50}: 1.00$}
\label{groundtruth}
\end{subfigure}
\hfill
\begin{subfigure}[b]{0.24\textwidth}
\centerline{\includegraphics[width=\textwidth]{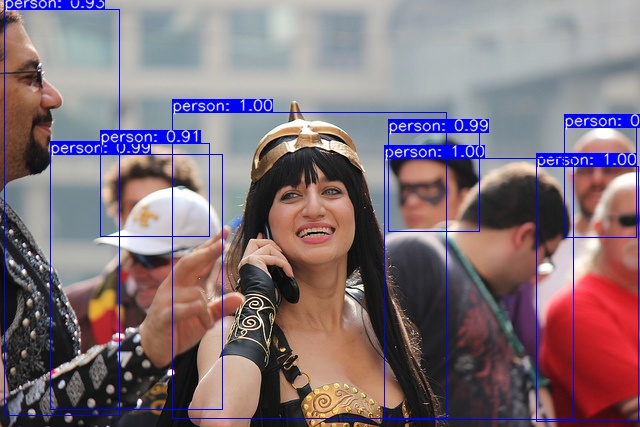}}
\caption{AWS, $AP_{50}: 0.64$}
\label{awssection2}
\end{subfigure}
\hfill
\begin{subfigure}[b]{0.24\textwidth}
\centerline{\includegraphics[width=\textwidth]{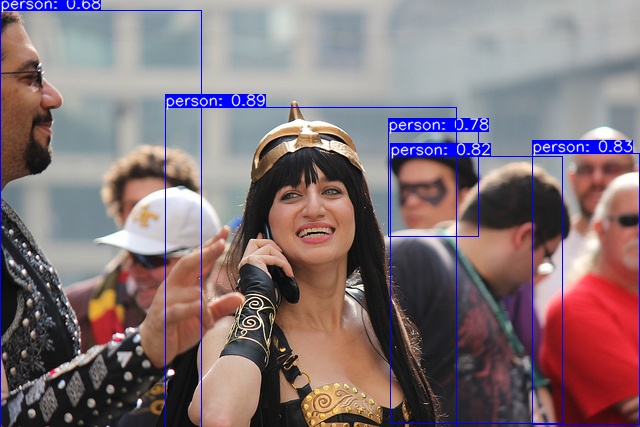}}
\caption{Azure, $AP_{50}: 0.56$}
\label{azuresection2}
\end{subfigure}
\hfill
\begin{subfigure}[b]{0.24\textwidth}
\centerline{\includegraphics[width=\textwidth]{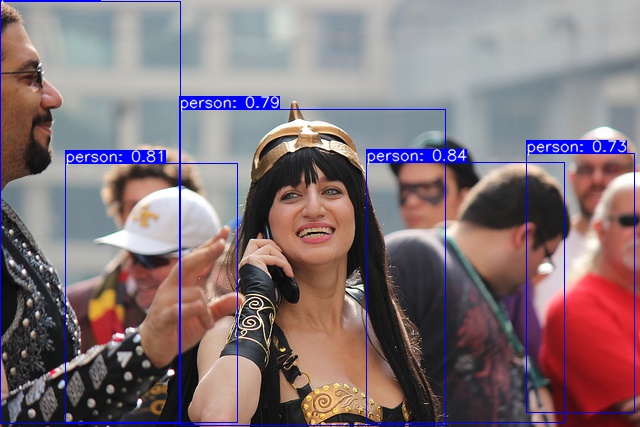}}
\caption{Google, $AP_{50}: 0.56$}
\label{googlesection2}
\end{subfigure}
\hfill
\begin{subfigure}[b]{0.24\textwidth}
\centerline{\includegraphics[width=\textwidth]{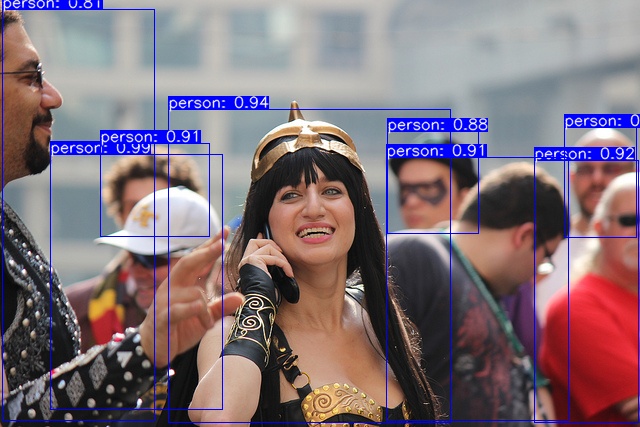}}
\caption{AWS+Azure, $AP_{50}: 0.71$}
\label{awsazuresection2}
\end{subfigure}
\hfill
\begin{subfigure}[b]{0.24\textwidth}
\centerline{\includegraphics[width=\textwidth]{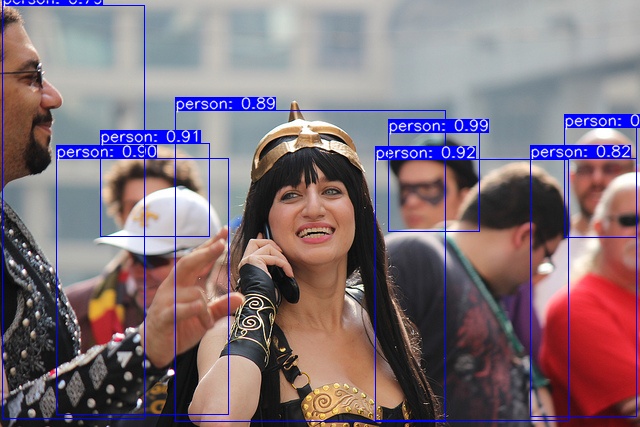}}
\caption{AWS+Google, $AP_{50}: 0.69$}
\label{awsgooglesection2}
\end{subfigure}
\hfill
\begin{subfigure}[b]{0.24\textwidth}
\centerline{\includegraphics[width=\textwidth]{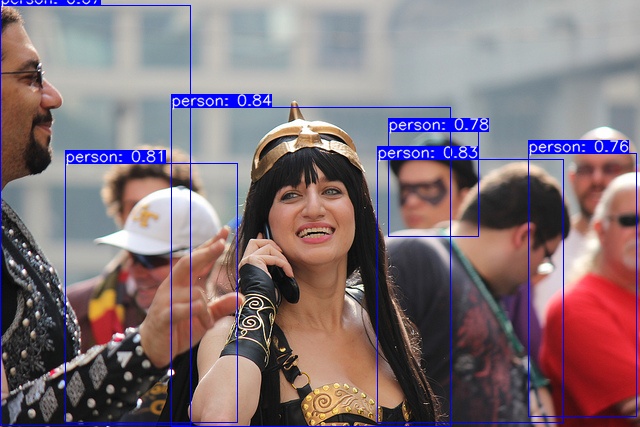}}
\caption{Azure+Google, $AP_{50}: 0.67$}
\label{azuregooglesection2}
\end{subfigure}
\hfill
\begin{subfigure}[b]{0.24\textwidth}
\centerline{\includegraphics[width=\textwidth]{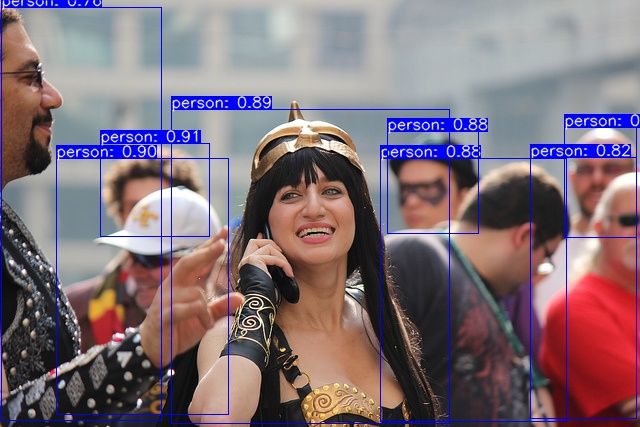}}
\caption{Three providers, $AP_{50}: 0.68$}
\label{awsazuregooglesection2}
\end{subfigure}
\caption{Detections and $AP_{50}$ of different cloud combinations.}
\label{fig:benifitsfed}
\end{figure*}

\subsection{Why We Need MLaaS Federation}

In Tab.~\ref{mapofdiffods}, we compare the $AP$ on predictions of the COCO Val $2017$ from AWS Rekognition, Azure Computer Vision, and Google Cloud Vision AI. We find that Azure has the worst performance on average. However, it does not mean that Azure performs poorly on every category in the dataset. We select top-$10$ frequent categories in COCO Val $2017$ and compare the $AP_{50}$ of the predictions from AWS, Azure, and Google on these categories. In Fig.~\ref{fig:top_10_cmp}, AWS is the best for categories such as ``person'', ``chair'', ``car'', and ``handbag''. Azure is the best for categories such as ``cup'', ``bottle'', and ``dining table'' while AWS did not identify any objects on these three categories. Google is the best on category ``book''. \emph{These phenomena indicate that for input with different features, the most appropriate MLaaS provider differs.} 

We next reveal the benefit of MLaaS federation. Following the ensemble strategies which will be introduced in Sec.~\ref{ensemble}, we have the $AP_{50}$ of AWS, Azure and Google are $0.64$, $0.56$ and $0.56$, and the $AP_{50}$ of the ensemble predictions from three MLaaS providers is $0.68$, as is demonstrated in Fig.~\ref{fig:benifitsfed}. We can see that the ensemble predictions from three MLaaS providers have higher $AP_{50}$ than the prediction from a single provider, verifying that the federation of MLaaS providers can provide more accurate prediction. In addition, by comparing Fig.~\ref{awsazuresection2} and Fig.~\ref{awsazuregooglesection2}, we find that the ensemble predictions of AWS and Azure ($AP_{50}=0.71$) is better than the ensemble predictions of three cloud providers ($AP_{50}=0.68$). 
\emph{These phenomena suggest that adding multiple MLaaS providers to inference can achieve higher accuracy than a single one. Still, more MLaaS providers added do not mean that we can gain higher accuracy.}

\subsection{Why MLaaS Federation is Possible}
By analyzing the TCP packets, we discover that the latency of a request consists of the transmission latency and the inference latency. The transmission latency is determined by the input data size and the round trip time (RTT) between the location of the client and the region of the MLaaS provider.
The inference latency is determined by the MLaaS itself, which is independent of the network conditions. Considering that the size of returned data is very small, the download time is negligible.

\begin{figure}[!t]
\centering
\begin{subfigure}[b]{0.24\textwidth}
\centerline{\includegraphics[width=\textwidth]{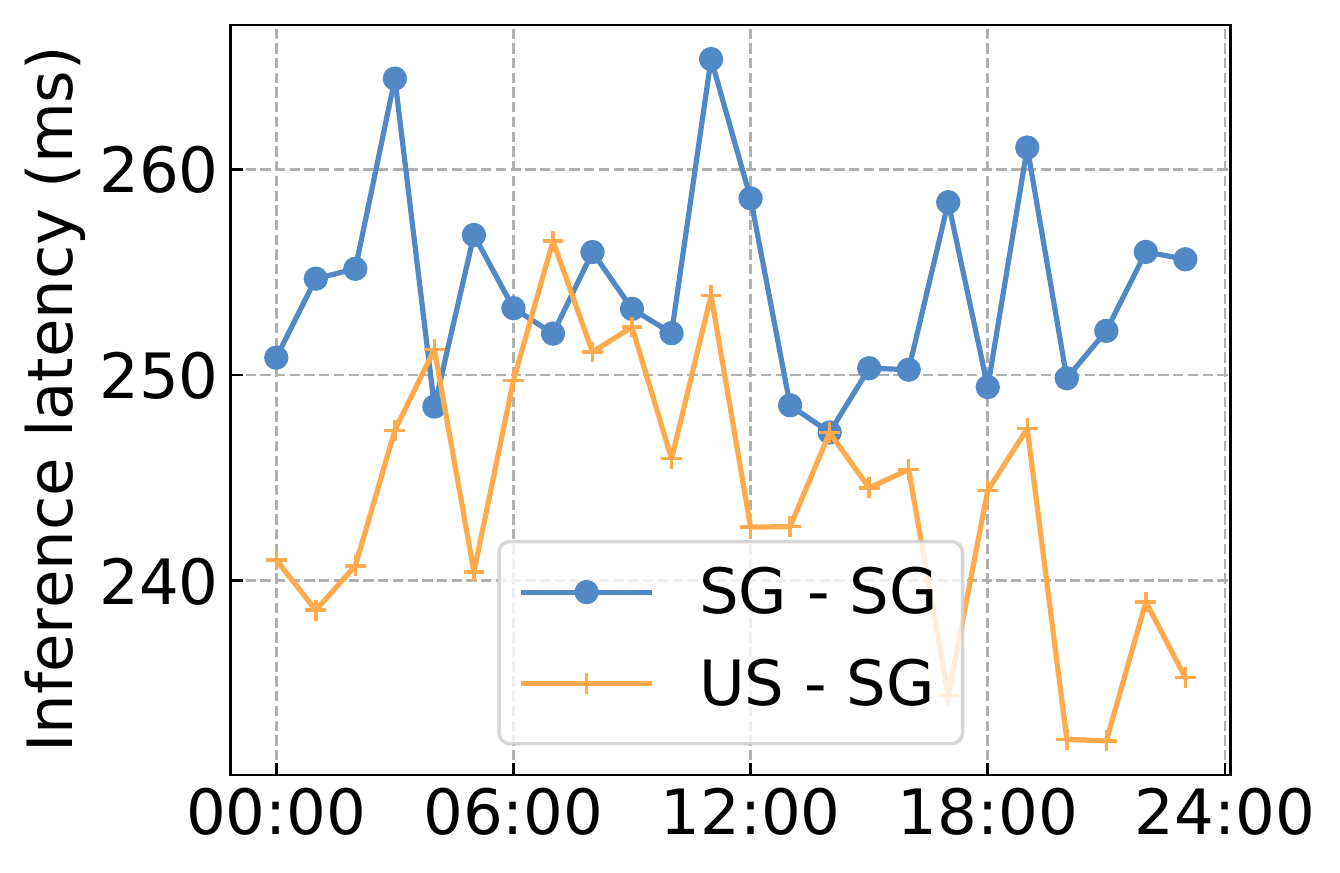}}
\caption{}
\label{confirmeawssglag}
\end{subfigure}
\hfill
\begin{subfigure}[b]{0.24\textwidth}
\includegraphics[width=\textwidth]{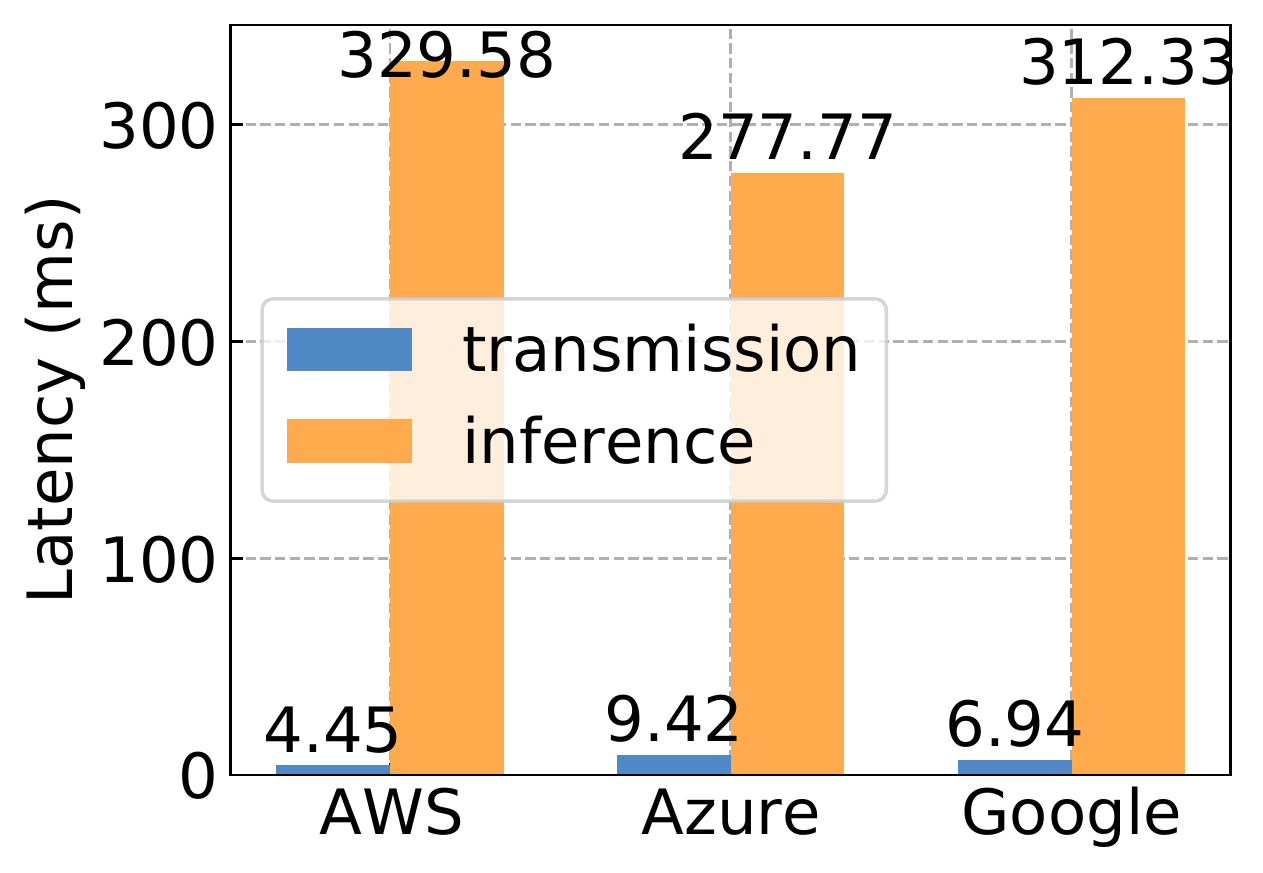}
\caption{}
\label{transmission_vs_inference}
\end{subfigure}
\caption{The inference latency in both routes (SG-SG and US-SG) are similar in $24$ hours, which indicates that the way we divide the total latency is  correct. The transmission latency is much less than inference latency.}
\label{fig:latency}
\end{figure}

We compare the inference latency parsed by our TCP packets in two routes, namely requesting MLaaS in Singapore from Singapore (SG-SG) and requesting MLaaS in Singapore from the US (US-SG). Both routes request MLaaSes within the region of Singapore, so theoretically, the inference latency should be similar. By analyzing the measurement results in Fig.~\ref{confirmeawssglag}, we find that the inference latency of both SG-SG and US-SG is similar within $24$ hours in a day, which proves that the way we divide the total latency is correct.
 
Since the user must send requests to multiple MLaaS providers, we consider the latency in such a case. The user device transfers input data to various MLaaSes by HTTPS, which indicates that the user device sends $n$ inputs to the MLaaSes sequentially via the same route in the transmission phase. Thus, the transmission latency is equal to the sum of the time to send the $n$ inputs. In the inference phase, $n$ MLaaSes predict the results in parallel, so the inference latency equals the maximum of the inference time among the $n$ MLaaSes. In Fig.~\ref{transmission_vs_inference}, we find that the transmission latency is much smaller than the inference latency. With sufficient network bandwidth, the above phenomenon indicates that although our transmission latency increases linearly with the number of MLaaS providers, the total latency will not increase linearly with MLaaS providers.

\section{System Model and Problem Formulation} \label{problemformulation}

MLaaS can be seen as a function that gets input data, such as an image, a text or a speech, and returns the prediction of the input, such as image category, translated text or text generated by the input speech. The specific forms of the input and output depend on the task targeted by MLaaS. Here we generalize it and assume that there are a set of inputs $\mathbf{I} = \{I_1, I_2, ..., I_T\}$ to be processed by $N$ available MLaaS providers. We denote $\mathbf{a}_t = [a_{t,1}, a_{t,2},...,a_{t,N}], a_{t, i} \in \{0, 1\}, i \in \{1, ..., N\}$ as the combination of selected MLaaS providers, i.e., $\mathbf{a}_t \in \{0, 1\}^N$. We denote $c_{t,i}$ as the cost to request the $i$-th MLaaS provider $M_i$ at timestamp $t$. Then the cost of the combination $\mathbf{a}_t$ can be denoted as $c_t = \sum_{i = 1}^{N}{c_{t, i}a_{t,i}}$. We denote $P_{\mathbf{a}_t}$ as the ensemble prediction from selected MLaaS providers determined by action $\mathbf{a}_t$. Then the final predictions for all input data can be represented as $\mathcal{P} = [P_{\mathbf{a}_1}, P_{\mathbf{a}_2}, ..., P_{\mathbf{a}_T}]$. We denote $v_{\mathbf{a}_t}$ to represent the accuracy of the prediction $P_{\mathbf{a}_t}$. 

We strive to identify the appropriate selection of MLaaS providers to maximize the accuracy and minimize the cost for all input data. In summary, the MLaaS federation problem ($\Omega$) can then be formally formulated as:



\begin{align}
& \max \ \ \ \ \ \ \ \ \sum_{t \in T} (v_t + \beta c_t), \\
& \begin{array}{ccc}
s.t. & \mathcal{F}(\mathcal{P})  \geq A_o, \\
     & \sum_{t=1}^{T}{\sum_{i=1}^{N}{c_{t, i}a_{t,i}}}  \leq C_o, \\
     & \sum_{i=1}^{N}{a_{t, i}} \neq 0, t \in \{1, 2, ..., T\}, \\
     & a_{t, i} \in \{0, 1\}, \\
     & \mathcal{P} = [P_{\mathbf{a}_1}, P_{\mathbf{a}_2}, ..., P_{\mathbf{a}_T}],
\end{array}
\end{align} 
where $\beta$, usually non-positive, is a hyperparameter to adjust the preference between accuracy and cloud cost, $\mathcal{F}$ is a function mapping from the predictions of all input data to accuracy, $A_o$ is the target accuracy, and $C_o$ is the overall budget for processing the whole workload.

The MLaaS federation problem $\Omega$ is NP-complete. The goal of $\Omega$ is to maximize the accuracy while minimizing the cost. Assume we only consider the computation of the provider selection part, and we have $T$ inputs and $N$ MLaaS providers. Then we have to predict whether the combination of $N$ MLaaS providers is optimal for each input or not. Even if we know the ground truth for each input, we need $O(2^N)$ comparisons to know the optimal combination corresponding to this input. Thus, the overall time complexity is $O(T\cdot 2^N)$. 

In a practical application environment, the prior information of input and corresponding optimal MLaaS provider combination is seldom available. Model-based solutions may not sufficiently adapt to the request dynamics and make intelligent provider selection decisions. In addition, the model and the dataset used to train the model of MLaaS are updating with the evolution of deep learning algorithms, making it difficult to achieve global optimally, especially in considering a long-term optimization.

The recent success in combinatorial RL provides an alternative perspective for this problem. Combinatorial RL reduces the complexity of the provider selection part. The rich historical viewer request patterns offer invaluable data resources that could be utilized for a data-driven provider selection problem. Specifically, the learning-based approach can not only well capture the hidden dynamics of input data and the model behind MLaaS but also enable an end-to-end solution from input data to MLaaSes' combination decision. Given these unique advantages, we present a combinatorial RL-based approach to solving this problem in the next section.

\section{MLaaS Federation Framework} \label{systemdesign}
\begin{figure*}[!t]
\centerline{\includegraphics[width=0.99\textwidth]{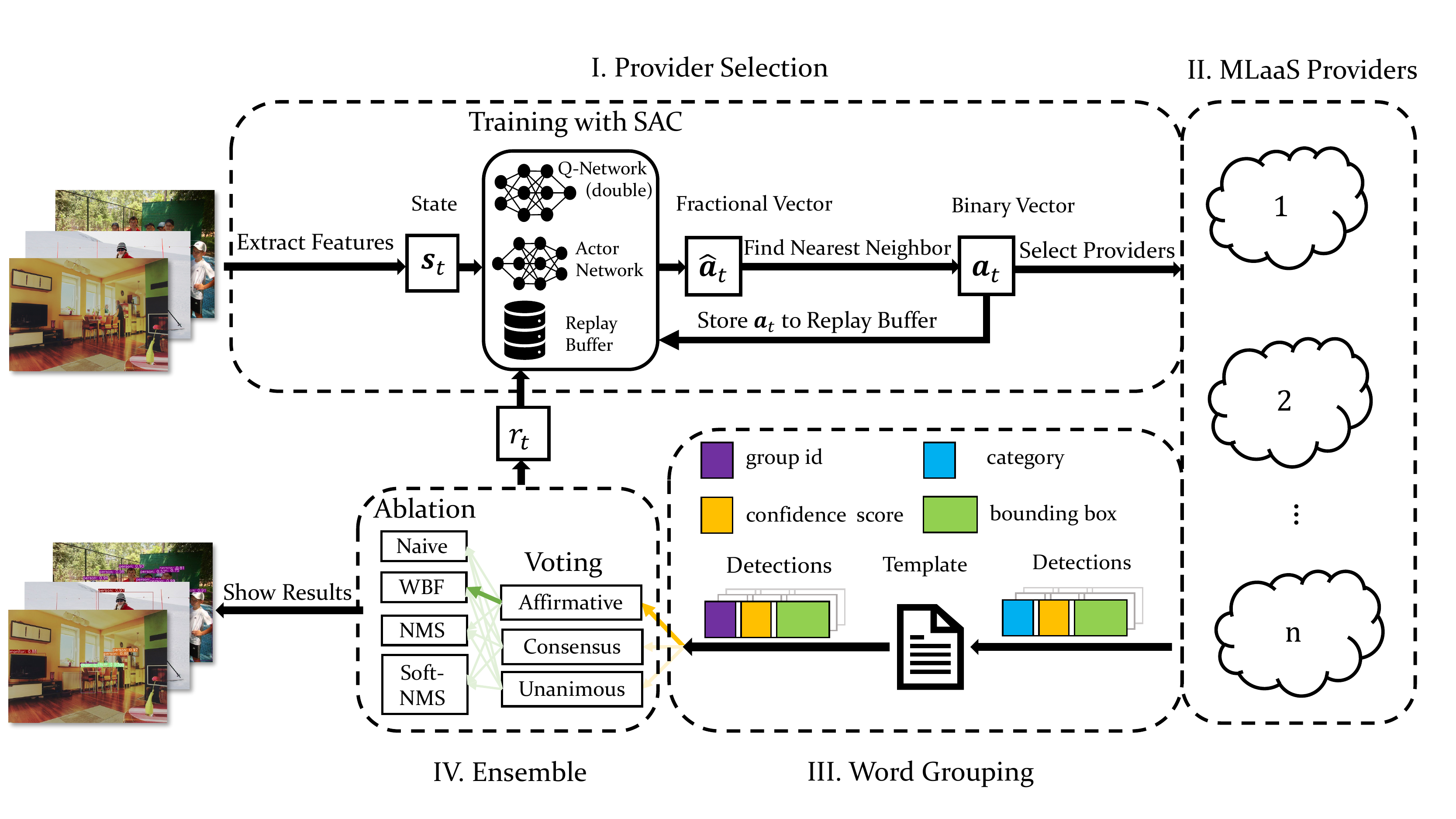}}
\caption{The overview of Armol. For the object detection task, we need to group the category names with the same meaning into exact representation in the word grouping part; we need to ablate the redundant detections in the ensemble part.}
\label{frameworkoverview}
\end{figure*}

In this section, we present Armol, a combinatorial RL-based cost-effective MLaaS federation framework that adaptively makes decisions about which providers to request to maximize the accuracy while minimizing the cost. We start by introducing the workflow of our framework. We then present the details of the provider selection part, the word grouping part, and the ensemble part.

\subsection{Framework Overview}

We consider a typical edge scenario for object detection. As shown in Fig.~\ref{frameworkoverview}, Armol receives an image at the beginning. In the provider selection part, we first extract the image features (i.e., state $\mathbf{s_t}$) at the edge-side client, and then we generate the proto action $\mathbf{\hat{a}_t}$ based on the image features by using the actor-network trained on the soft actor-critic (SAC) algorithm \cite{haarnoja2018soft}. However, the proto action is a fractional vector, we have to map it to a binary vector $\mathbf{a_t}$. Then the edge client requests the providers selected by action $\mathbf{a_t}$ and waits until receiving all the predictions from selected providers. Cloud $1$, $2$ to $n$ are the available MLaaS providers. Next is the word grouping part. Since different cloud services may use different words to represent the same category, we need this part to identify and unify words with the same meaning into one form to ensure that the subsequent ensemble part can be performed correctly. Finally, there is the ensemble part, which aims to ablate duplicate predictions while retaining the correct ones. In total, there are $12$ pathways to choose from, and we end up with the Affirmative-WBF path based on our measurements in Sec.~\ref{measurement}. Armol also generates reward $r_t$ in the ensemble part, which will be store in replay buffer with the binary action vector $\mathbf{a_t}$, the state $\mathbf{s_t}$ and the next state $\mathbf{s_{t+1}}$. Only after going through the above modules, the final prediction can be drawn on the image. The remainder of this section covers the details of Armol.

\subsection{Provider Selection: A Combinatorial RL Approach}
We first describe the design of state, action, and reward for our MLaaS federation problem.

\textbf{State.} To facilitate the training of the model, we use a pre-trained MobileNet, a classical model on the image classification task, to extract the  feature of input, which represents the state $\mathbf{s}_t$ obtained from the environment at timestamp $t$. This method is described on the left top of Fig.~\ref{frameworkoverview}.

\textbf{Action.} A subset of $N$ MLaaS providers can be represented by a vector $\mathbf{a}_t \in \mathcal{A} = \{0, 1\}^N$ and $\mathbf{a}_t \neq \{0\}^N$, where the $i$-th element $1$ means that the $i$-th provider is in this subset, while $0$ means not. i.e, $\mathbf{a}_t = [a_{1,t},...,a_{N,t}], a_{i,t} \in \{0, 1\}$. If we have $N$ available MLaaS providers, then the size of action set $\mathcal{A}$ is $2^N-1$. Thus, the action space of our MLaaS federation problem is an exponential multiple of $N$. When $N$ is large, it is hard for RL algorithms with discrete action spaces to handle the action spaces with size $2^N - 1$. Thus, we have to solve this problem by mapping the $\mathbf{\hat{a}}$ from continuous action spaces to an element in discrete binary vector set $\mathcal{A}$:
\begin{gather} 
    \tau: \mathbb{R}^n \rightarrow \mathcal{A}, \\
    \tau(\mathbf{\hat{a}}) = \mathop{\arg\min}_{\mathbf{a} \in \mathcal{A}}{|\mathbf{a} - \mathbf{\hat{a}}|^2},
\end{gather}
where $\tau$ is the nearest-neighbor mapping from a continuous space $\mathbb{R}^n$ to the discrete binary vector set $\mathcal{A}$. It returns the action $\mathbf{a}$ that is closest to $\mathbf{\hat{a}}$ by $l_2$ distance. The action $\mathbf{a}$ will be stored to replay buffer later with other elements.

\textbf{Reward.} There are two modes of the training process, offline and online, and the definition of rewards in the two modes are different. As described in \cite{mapvoc2012}, we need the ground truth to calculate the mAP. However, in practical applications, not all input images have ground truth. For images without ground truth, we use the ensemble prediction of $N$ MLaaS providers as the ground truth. As shown in Fig.~\ref{fig:benifitsfed}, although the mAP of $N$-providers ensemble prediction is not optimal, it is still better than the prediction of a single provider. Thus, it is feasible to use the ensemble prediction of all available MLaaS providers as ground truth. In addition to increasing the mAP, we also want to use as few MLaaS providers as possible to reduce the inference fee. In summary, the reward can be defined as follows:
\begin{equation}
    r_t = v_t + \beta c_t,
\end{equation}
where $v_t$ is the $AP_{50}$ of prediction, $c_t$ is the cost to request the subset of MLaaS providers selected by action $\mathbf{a}_t$, $\beta$ is a hyperparameter, usually a non-positive number, to ensure that the action with a lower cost is selected. It is possible that providers selected by action $\mathbf{a}$ will not return any prediction, for which case we define the reward as $-1$.

We leverage SAC to train the RL agent. SAC is an off-policy actor-critic algorithm based on the maximum entropy RL framework. \cite{haarnoja2018soft} explains the principle of SAC. Thus, we describe the details of training the RL agent next.

The algorithm of training the RL agent is proposed in Algo.~\ref{alg:traingsac}. First, we initialize the replay buffer $\mathcal{B}$, the hyperparameter $\beta$, and the parameters for two Q-networks, two target Q-networks, and an actor-network. We use a fully connected network (FCN) with two hidden layers to represent the above networks, and the difference between the Q-network and actor-network is the input and output layers. Our training algorithm makes use of two soft Q-functions to mitigate positive bias in the policy improvement step that is known to degrade the performance of value-based methods \cite{fujimoto2018addressing}. 

Second, for each step, we observe the input image and extract the feature as the state. We select the action $\mathbf{\hat{a}}$ by policy $\pi_{\theta}(\cdot \vert \mathbf{s})$, then map $\mathbf{\hat{a}}$ to a binary action $\mathbf{a}$ and execute $\mathbf{a}$ in the environment to observe the next state $\mathbf{s}^{\prime}$, reward $r$, and done signal $d$. We next store $(\mathbf{s}, \mathbf{a}, r, \mathbf{s}^{\prime}, d)$ to replay buffer $\mathcal{B}$. 

Finally, if it is the step to update the networks, we sample a batch of transitions from replay buffer $\mathcal{B}$. The target of Q-network is given by:
\begin{multline}
\label{eq:target}
    y(r, \mathbf{s}^{\prime}, d) = \\ r +  \gamma(1-d)\left(\min_{j=1,2} Q_{\phi_{targ,j}}(\mathbf{s}^{\prime}, \mathbf{\tilde{a}}^{\prime}) - \alpha \log\pi_{\theta}(\mathbf{\tilde{a}}^{\prime} \vert \mathbf{s}^{\prime}) \right),
\end{multline}
where $\mathbf{\tilde{a}}^{\prime}$ is sampled from $\pi_{\theta}$:
\begin{equation}
     \mathbf{\tilde{a}}^{\prime} \sim \pi_{\theta}( \cdot \vert \mathbf{s}^{\prime}).
\end{equation}
Then we can update two Q-networks $Q_{\phi_i}, i = 1, 2$ by one step of gradient descent using:
\begin{equation}
\label{eq:updateq}
    \nabla_{\phi_i}\frac{1}{|B|}\sum_{(\mathbf{s}, \mathbf{a}, r, \mathbf{s}^{\prime}, d) \in B}(Q_{\phi_{i}}(\mathbf{s},\mathbf{a}) - y(r, \mathbf{s}^{\prime}, d))^2,
\end{equation}
and update the policy by one step of gradient ascent using:
\begin{equation}
\label{eq:updatepolicy}
    \nabla_{\theta}\frac{1}{|B|}\sum_{\mathbf{s} \in B}\left(\min_{i=1,2}{Q_{\phi_{i}}(\mathbf{s},\mathbf{\tilde{a}}_{\theta}(\mathbf{s}))} - \alpha\log\pi_{\theta}(\mathbf{\tilde{a}}_{\theta}(\mathbf{s}) \vert \mathbf{s}) \right),
\end{equation}
where $\mathbf{\tilde{a}}_{\theta}(\mathbf{s})$ is a sample from $\pi_{\theta}( \cdot \vert \mathbf{s}^{\prime})$.
At last we update the target networks $Q_{targ, i}, i = 1, 2$ with:
\begin{equation}
\label{eq:updatetarget}
\phi_{targ, i} \leftarrow \rho\phi_{targ, i} + (1 - \rho) \phi_{i}.
\end{equation}

Note that our implementation of SAC omits the extra value function because Q-function and value function can represent each other \cite{haarnoja2018soft}. This part is in the top center of the Fig.~\ref{frameworkoverview}, Armol will request the MLaaS providers selected by the action $\mathbf{a}_t$ next.


\begin{algorithm}
\caption{Training the RL agent with SAC}
\label{alg:traingsac}
\SetAlgoLined
Initialize policy parameters $\theta$;\\
Initialize Q-function parameters $\phi_1, \phi_2$; \\
Initialize replay buffer $\mathcal{B}$; \\
Initialize hyperparameter $\beta$; \\ 
Set target Q-function parameters equal to main parameters $\phi_{targ, 1} \leftarrow \phi_1, \phi_{targ, 2} \leftarrow \phi_2$;\\
\For{time=1,...}{
    Observe an input image, extract state $\mathbf{s}$ and select action $\hat{\mathbf{a}} \sim \pi_{\phi}(\cdot \vert \mathbf{s})$; \\
    Get the nearest binary vector $\mathbf{a}$ of $\hat{\mathbf{a}}$ in $l_2$ distance; \\
    Request the MLaaS providers selected by $\mathbf{a}$; \\
    Store $(\mathbf{s}, \mathbf{a}, r, \mathbf{s}^{\prime}, d)$ to replay buffer $\mathcal{B}$; \\
    \If{it's time to update}{
        \For{j in range(update times)}{
            Randomly sample a batch of transitions $B = \{(\mathbf{s}, \mathbf{a}, r, \mathbf{s}^{\prime}, d)\}$ from $\mathcal{B}$; \\
            Compute targets for the $Q_{\phi_1}, Q_{\phi_2}$ using Eq.~\ref{eq:target}; \\
            Update $Q_{\phi_1}, Q_{\phi_2}$ by one step of gradient descent using Eq.~\ref{eq:updateq}; \\
            Update policy by one step of gradient ascent using Eq.~\ref{eq:updatepolicy};
            \\
            Update target Q-networks using Eq.~\ref{eq:updatetarget}; \\
        }
    }
}
\end{algorithm}


\subsection{Word Grouping Part} \label{wordgroupalgo}
After receiving the predictions from the selected providers, we need to standardize the presentation of the returned predictions to prevent ambiguity. This part is task-oriented since different tasks have different outputs. Here we take object detection, a classical computer vision task, as an example. 

As we mentioned in Sec.~\ref{problemformulation}, an object detection cloud service can be seen as a function that returns a list of detections $D = [d_1, d_2, ..., d_{l_d}]$ where $d_i$ is given by a triple $[l_i, f_i, b_i]$ that consists of the corresponding category $l_i$, the corresponding confidence score $f_i$, and a bounding box $b_i$. The length of detection list $D$, namely $l_d$, represents the number of objects detected on this image. However, the different services may return the different category names for the object in the same category, i.e., ``motorbike'' vs. ``motorcycle''. It is clear that these category names have the same meaning, so we propose a feasible algorithm to aggregate these category names into one group.

The following are the details. First of all, the user must provide a template $T$ that contains all the category names they need. Here we use the $80$ categories of the COCO dataset as $T$ and find the close synonyms of category names in $T$ based on the synonyms dataset extracted from WordNet. Subsequently, based on the measurement results, we collect the category names from all MLaaS providers as set $A$. However, we find that the synonyms from WordNet are not enough to cover all words in $A$, so we manually add the missing words within set $A$ to the $80$ groups. After that, we discard the remaining words in set $A$ that are irrelevant to the $80$ categories in the COCO dataset. For words in the same group, we consider they have the same meaning when used as nouns. Finally, a single detection $\mathbf{d}_i$ can be given by a triple $[n_i, f_i, b_i]$, where $n_i$ is the group index of the corresponding category $l_i$. Only with this problem solved can we compare the accuracy of different MLaaS providers and ensemble the predictions correctly. This part is on the central bottom of Fig.~\ref{frameworkoverview}, where the client has received the predictions from the selected MLaaS providers.

\subsection{Ensemble Part} \label{ensemble}
This ensemble part is also task-oriented. Fig.~\ref{fig:benifitsfed} illustrates the mAP of different combinations of three providers. We find that a single MLaaS provider has low mAP while the ensemble of multiple MLaaS providers reaches excellent performance. Therefore, based on the measurement results in Sec.~\ref{measurement}, we propose a novel strategy to ensemble the predictions from different object detection service providers. We divide this part into two steps: voting and ablation.

\textbf{Voting methods.} Common voting methods include ``Affirmative", ``Consensus" and ``Unanimous" \cite{DBLP:conf/ecai/Casado-GarciaH20}. To conduct the voting methods, we need to standardize the presentation of the predictions from different MLaaS providers first. Then we group the detections of the image into $G = [g_1, g_2, ..., g_r]$, where $g_i, i \in \{1, 2,..., r\}$ is a list of detections and $r$ represents the total number of objects detected by $N$ cloud service providers. For detections $d_{p}, d_{q} \in g_i$, they must confirm that $IoU(b_p, b_q) > 0.5$ and $n_p = n_q$. $IoU(a, b)$ is calculated by dividing the area of intersection between box $a$ and box $b$ by the area of union between $a$ and $b$. Then we adjust the three voting methods to our work, which are described as follows.
\begin{itemize}
    \item Affirmative. This method keeps all groups in $G$, which means that the detection is valid whenever one of the clouds says that a region contains an object.
    \item Consensus. This method holds the groups with a size greater than $N/2$, meaning most clouds must agree that a region contains an object.
    \item Unanimous. Only the groups whose size is equal to $N$ are kept in this method, which means that all the object detection cloud services must agree to consider that a region contains an object.
\end{itemize}

We choose affirmative as the primary voting method. Because the three MLaaS providers are more of a complementary relationship and their predictions do not overlap much, as analyzed in Sec.~\ref{measurement}. Therefore, consensus or unanimous methods may remove some of the true-positive results. Besides, the evaluation results from \cite{DBLP:conf/ecai/Casado-GarciaH20} on different models also indicate that the affirmative method is superior to other methods.

\textbf{Ablation Methods.} Boxes in a group may repeatedly express an object, increasing the number of false-positive predictions and leading to a lower mAP. To reduce useless boxes (i.e., false-positive predictions), Non-Maximum Suppression (NMS) \cite{hosang2017learning}, Soft-NMS \cite{bodla2017soft}, and Weighted Boxes Fusion (WBF) \cite{solovyev2019weighted} are proposed. NMS only saves the box with the most significant confidence score among a group and discards all other boxes. However, NMS inevitably removes the detections of some highly overlapped objects. Thus, Bodla et al. propose Soft-NMS to solve this problem. Instead of completely removing the detections with high IoU, it reduces the confidence score of the detections proportional to the IoU value. However, both NMS and Soft-NMS discard redundant boxes and thus can not effectively produce averaged localization predictions from different models. WBF takes the weighted average of the box coordinates within a group as the retained box, where the weight is the confidence score of the boxes. Moreover, the confidence score of the retained box is the average of the confidence score of boxes within a group.

\begin{figure}[!t]
\centering
\includegraphics[width=0.35\textwidth]{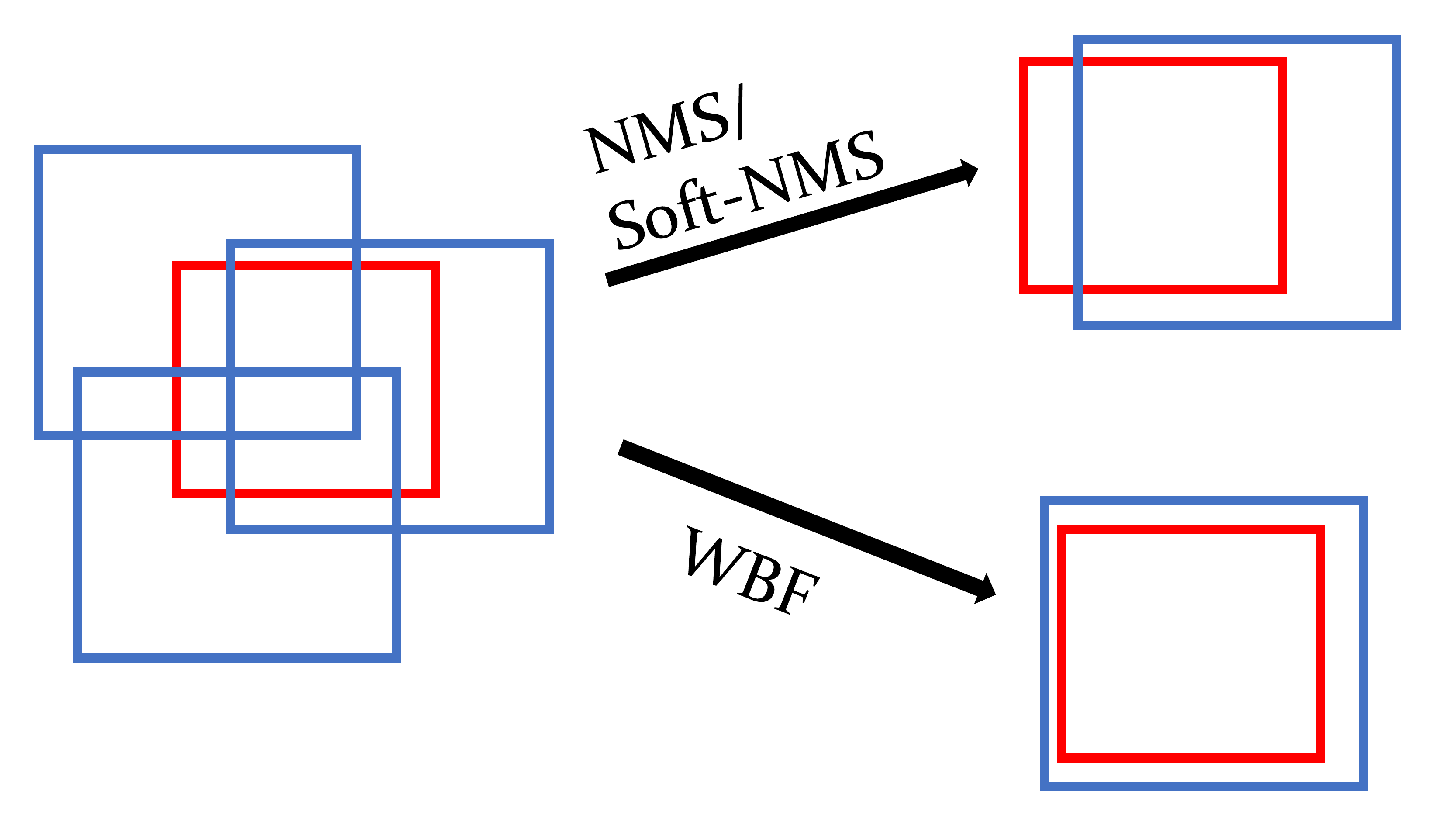}
\caption{Red boxes represent the ground truth. The left blue boxes represent the detections from AWS, Azure, and Google. The right top blue box is the box kept by NMS or Soft-NMS; the right bottom blue box is the box generated by WBF \cite{solovyev2019weighted}.}
\label{fig:ablation}
\end{figure}

Our measurements in Sec.~\ref{measurement} find that the differences between AWS, Azure, and Google are relatively significant. As shown in Fig.~\ref{fig:ablation}, for the same object, the boxes of all three cloud services are inaccurate and scattered in three directions. If we use NMS or Soft-NMS methods to ablate the boxes, the box kept is still inaccurate. However, WBF can obtain a more accurate predicted box by fusing the boxes in all three directions. Therefore, we decide to use the WBF method to ablate the duplicated boxes.

As described in Fig.~\ref{frameworkoverview}, in the ensemble part, we first go through the voting method for each group of similar boxes and then remove the duplicate boxes by the ablation method.

\section{Performance Evaluation} \label{evaluation}
In this section, we conduct extensive experiments to evaluate the performance of the MLaaS provider selection part in Armol. Specifically, using real trace-driven evaluations, we demonstrate that the superiority of Armol over other benchmark approaches.

\subsection{Setup and Methodology}
\textbf{Evaluation Setup.} We collect the predictions of COCO Val $2017$ from AWS Rekognition, Azure Computer Vision, and Google Cloud Platform Vision AI as the environment. We open-source the code and data in \texttt{https://github.com/ShuzhaoXie/Armol}. The inference cost for AWS and Google Cloud Vision AI is $0.001$ USD per image, while the inference cost for Azure varies by about $10\%$ with the region. Our measurements were done in Singapore, and Azure's price in this region is $0.001$ USD. So in the following experiments, we set the inference cost of each request to an MLaaS as $0.001$ USD. 
We implement the algorithm of the RL agent based on the SpinningUp \cite{SpinningUp2018} framework and add support for GPU training, which runs on a server with an NVIDIA $1080$ Ti GPU card, an Intel(R) Xeon(R) CPU E$5$-$2650$ v4@$2.20$GHz, and $64$ GB memory. The RL environment is implemented in Python for compatibility. The learning rate $\eta$ for the actor-network and the Q-networks is $0.0001$, respectively. We set $\gamma$ as $0.9$, and $\alpha$ as $0.2$. In order to gain the best mAP, we set $\beta$ as $0$. The batch size is $1000$. The training epoch is $100$, and the steps per epoch are $2000$.

\textbf{Baseline Methods.} We compare our approach with several baseline methods as follows. 
\begin{itemize}
    \item \textbf{Random-1}: This method only gives a random selection of MLaaS providers for each image.
    \item \textbf{Random-N}: This method chooses a subset of available MLaaS providers for each image randomly.
    \item \textbf{Ensemble-N}: This method aggregates the predictions of all MLaaS providers. 
    \item \textbf{Armol-w/o gt}: This method uses the ensemble predictions of all MLaaS providers as the ground truth to generate the reward. The hyperparameter $\beta$ is set as $-0.1$ to select the action with the lower cost of inference fee. The other hyperparameters are the same as Armol.
    \item \textbf{Armol-P}: This method means that we train the RL agent on proximal policy optimization (PPO)  \cite{schulman2017proximal}, which is a classical on-policy RL algorithm that is worth comparing.
    \item \textbf{Armol-T}: This method means that we train the RL agent on twin delayed deep deterministic policy gradient (TD3) \cite{fujimoto2018addressing}. TD3 is a classical deterministic policy training algorithm, the comparison with which can demonstrate the benefits of the maximum entropy property of SAC.
    \item \textbf{Upper Bound}: To reach the goal of gaining more mAP while spending less money, based on the measurements in Sec.~\ref{measurement}, we use a brute-force search algorithm to select the best combination, which is demonstrated in Algo.~\ref{alg:brute}. The voting method is affirmative, and the boxes ablation method is WBF.
\end{itemize}

\begin{algorithm}
\caption{Brute Force Search Algorithm}
\label{alg:brute}
\SetAlgoLined
Initialize set $\mathcal{D}$ to store the best detection of all images;\\
\For{image $I_t$ in $\mathbf{I}$}{
    Initialize the max mAP $v_{max} = -1$; \\
    \For{action $\mathbf{a}$ in $\{0, 1\}^N-\{0\}^N$}{
        Get and ensemble the detection $D_{\mathbf{a}}$ by action; \\
        Calculate the mAP $v_{D_{\mathbf{a}}}$ of detections $D_{\mathbf{a}}$; \\
        \If{$v \geq v_{max}$}{
            Update $v_{max}$ to $v_{D_{\mathbf{a}}}$; \\
            Update the best action $\bar{\mathbf{a}}$ to $\mathbf{a}$; \\
            Update the best detection $\Bar{D_{\mathbf{a}}}$ to $D_{\mathbf{a}}$; \\
        }
    }
    $\mathcal{D} \leftarrow \mathcal{D} \cup \Bar{D_{\mathbf{a}}} $; \\
}
return $\mathcal{D}$. \\
 
\end{algorithm}

\textbf{Evaluation Metrics.} We use the following metrics:
\begin{itemize}
    \item \textbf{Cost}: We denote this metric as average cost $c_{e}$ in a test episode, in unit of $10^{-3}$ USD:
    \begin{equation}
        c_{e}= \frac{\sum_{t=0}^{T-1}c_t}{T}
    \end{equation}
    
    \item $\mathbf{AP_{50}}$: This metric means the average precision of predictions with a $50\%$ IoU threshold. We use $AP_{50}$ instead of $mAP$ is because we want to reduce the computation and speed up the training. Because $mAP$ is the average of APs with IoU threshold from $50\%$ to $95\%$ with a $5\%$ increase per step. $AP_{50}$ is the average precision with a $50\%$ IoU threshold, the computation of which is $10\%$ of mAP. Besides, $AP_{50}$ is also a standard metric in object detection tasks.
\end{itemize}

\subsection{Evaluation Results} 
To understand the performance of the provider selection part, we conduct experiments to demonstrate: 1) the superiority of training RL agent on SAC; 2) the feasibility to leverage predictions from all MlaaS providers as ground truth; 3) the scalability of our combinatorial RL approach. 

\begin{figure}[!t]
\centering
\begin{subfigure}[b]{0.24\textwidth}
\centerline{\includegraphics[width=\textwidth]{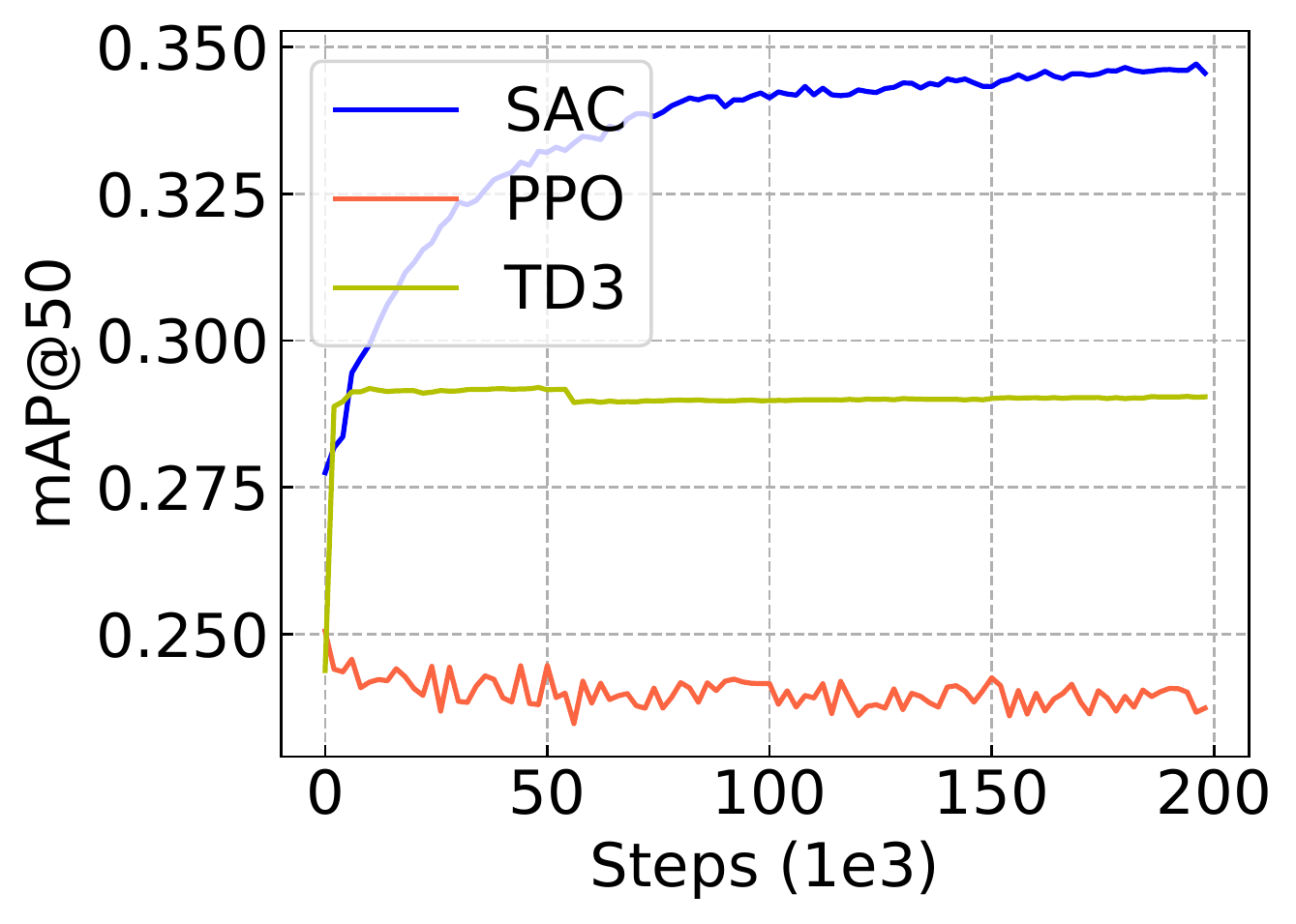}}
\caption{}
\label{mapsacppo}
\end{subfigure}
\hfill
\begin{subfigure}[b]{0.24\textwidth}
\includegraphics[width=\textwidth]{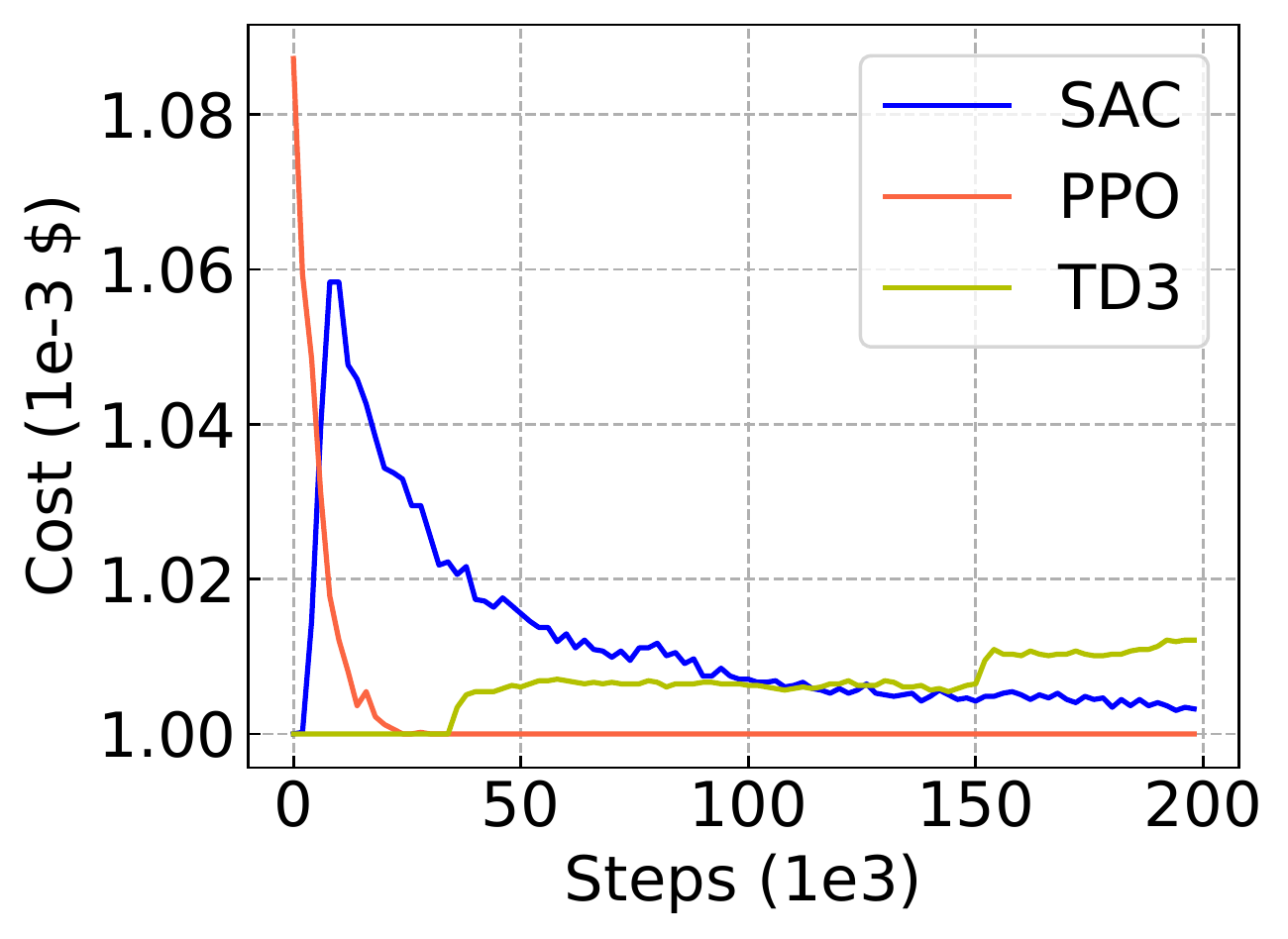}
\caption{}
\label{costsacppo}
\end{subfigure}
\caption{The label of the Y-axis on the left is $mAP@50$ of the whole episode, which is the same as $AP_{50}$. The label of the Y-axis on the right is the average cost per test episode. The training algorithm tests the RL agent at the end of every epoch. The figure shows the training process of Armol on SAC, Armol on PPO, and Armol on TD3.}
\label{fig:sacppo}
\end{figure}

\textbf{Superiority of training RL agent on SAC.} Tab.~\ref{pArmol_mtrc_2} shows the metric statistics of Armol on SAC, PPO and TD3. We can see that the mAP of Armol on SAC is greater than PPO and TD3, and the average cost is lower than TD3. In Fig.~\ref{fig:sacppo}, we can find that the SAC converges better and faster. Compared to Random-N, Armol on SAC has $3.09\%$ higher mAP and $41.75\%$ less inference fee. Compared to Ensemble-N, the Armol with ground truth (Armol-w/ gt) gains equal mAP and reduces $67\%$ inference fee.

\begin{figure}[!t]
\centering
\begin{subfigure}[b]{0.24\textwidth}
\centerline{\includegraphics[width=\textwidth]{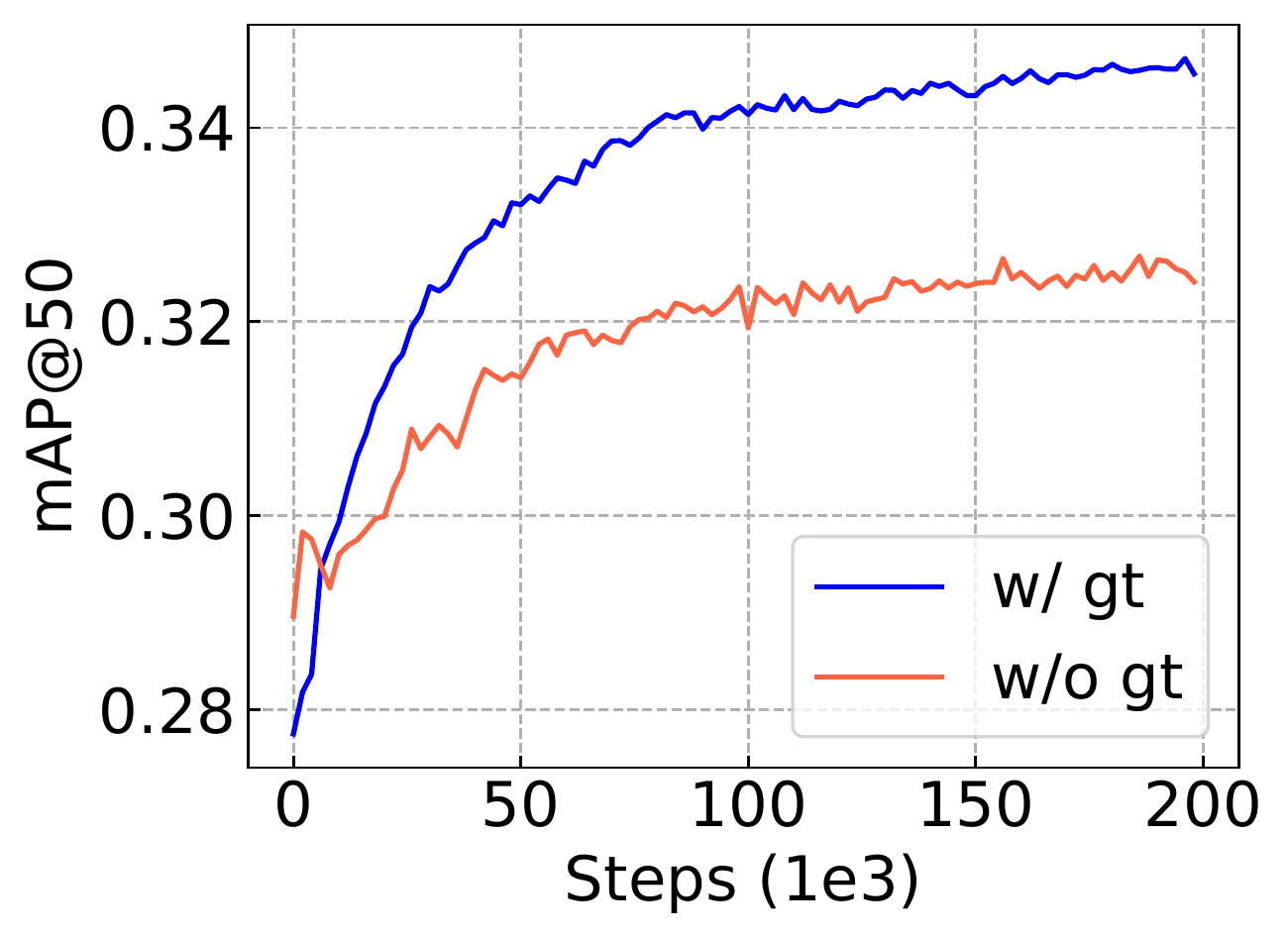}}
\caption{}
\label{mapwogt}
\end{subfigure}
\hfill
\begin{subfigure}[b]{0.24\textwidth}
\includegraphics[width=\textwidth]{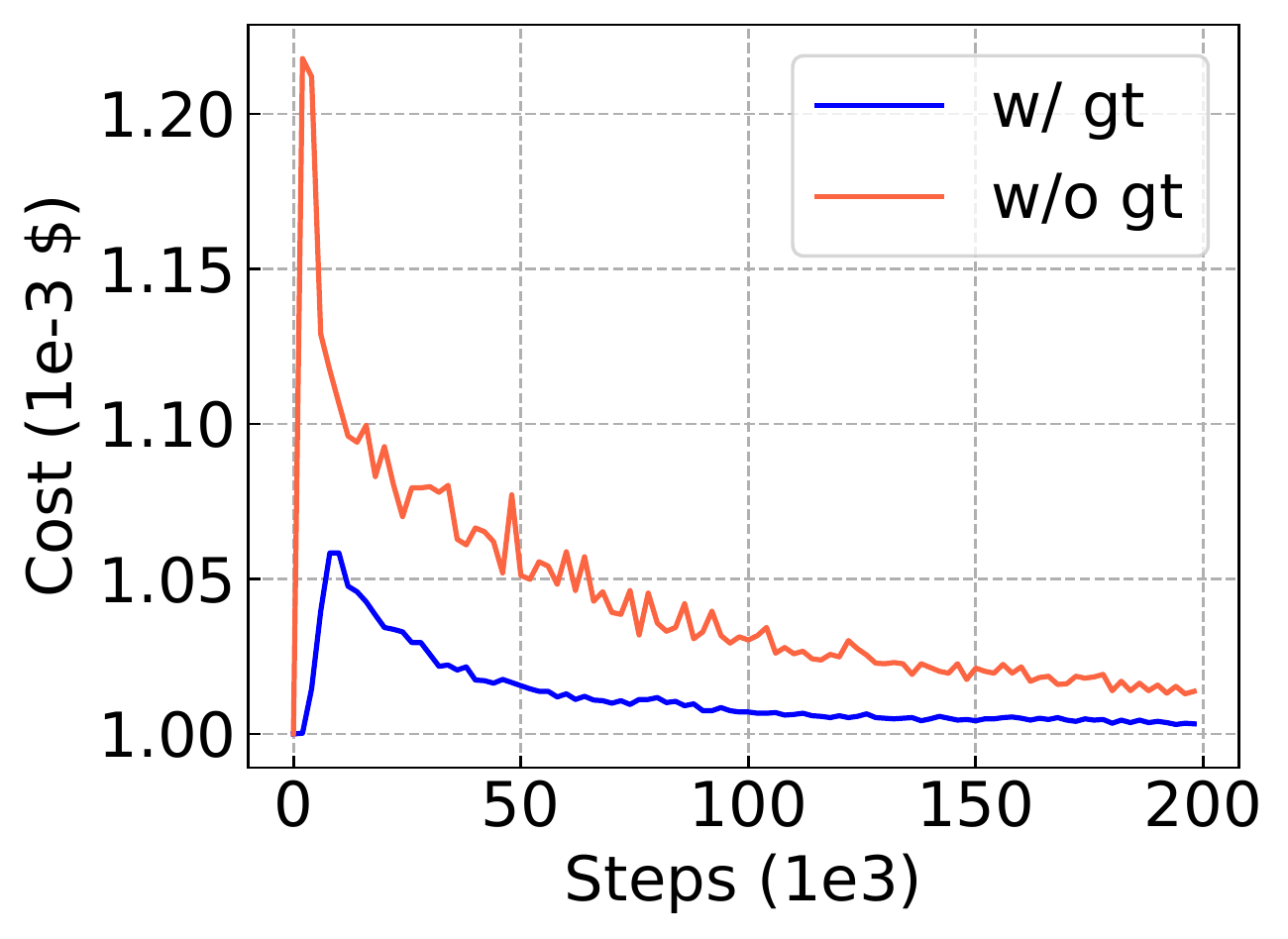}
\caption{}
\label{costwogt}
\end{subfigure}
\caption{w/ gt means the ``Armol with ground truth''; w/o gt means the ``Armol without ground truth''.}
\label{fig:wogt}
\end{figure}

\textbf{Feasibility to leverage predictions from three providers as ground truth.} Tab.~\ref{pArmol_mtrc_2} shows the without ground-truth method reduces $66\%$ cost compared to all federated predictions with only $4.3\%$ lower mAP. As can be seen in Fig.~\ref{fig:wogt}, Armol-w/o gt converges stably, although both $AP_{50}$ and cost are not as good as Armol-w/ gt in the training process.

\begin{table}[!t]
\caption{Performance metrics of different baselines. ``AWS'' means how many images in the test episode choose the AWS, while ``Azure'' and ``Google'' have the same meaning. ``Armol-w/ gt'' has the same meaning with ``Armol'' and ``Armol on SAC''. The unit of cost is $10^{-3}$ USD.}
\begin{center}
\begin{tabular}{|c|c|c|c|c|c|c|}
\hline
 \textbf{Methods} & \textbf{mAP} & $\mathbf{AP_{50}}$ & \textbf{Cost} & \textbf{AWS} & \textbf{Azure} & \textbf{Google}  \\ 
\hline
 Random-1 & $15.75$ & $24.49$ & $1.000$ & $1690$ & $1605$ & $1657$ \\  
\hline
 Random-N & $18.66$ & $28.89$ & $1.722$ & $2858$ & $2863$ & $2809$ \\
\hline
 Ensemble-N & $21.75$ & $34.69$ & $3.000$ & $4952$ & $4952$ & $4952$ \\
\hline
 \textbf{Armol-w/ gt} & $\mathbf{21.75}$ & $\mathbf{34.71}$ & $\mathbf{1.003}$ & $\mathbf{2863}$ & $\mathbf{950}$ & $\mathbf{1156}$ \\
\hline
 Armol-w/o gt & $20.81$ & $32.68$ & $1.016$ & $3426$ & $683$ & $924$ \\
\hline
 Armol-PPO & $14.99$ & $25.05$ & $1.087$ & $1300$ & $2541$ & $1543$ \\
\hline
 Armol-TD3 & $18.90$ & $29.20$ & $1.006$ & $4843$ & $114$ & $26$ \\
\hline
 Upper Bound & $23.83$ & $37.70$ & $1.202$ & $3881$ & $1126$ & $944$ \\
\hline

\end{tabular}
\label{pArmol_mtrc_2}
\end{center}
\end{table}

\begin{figure}[!t]
\centering
\begin{subfigure}[b]{0.24\textwidth}
\centerline{\includegraphics[width=\textwidth]{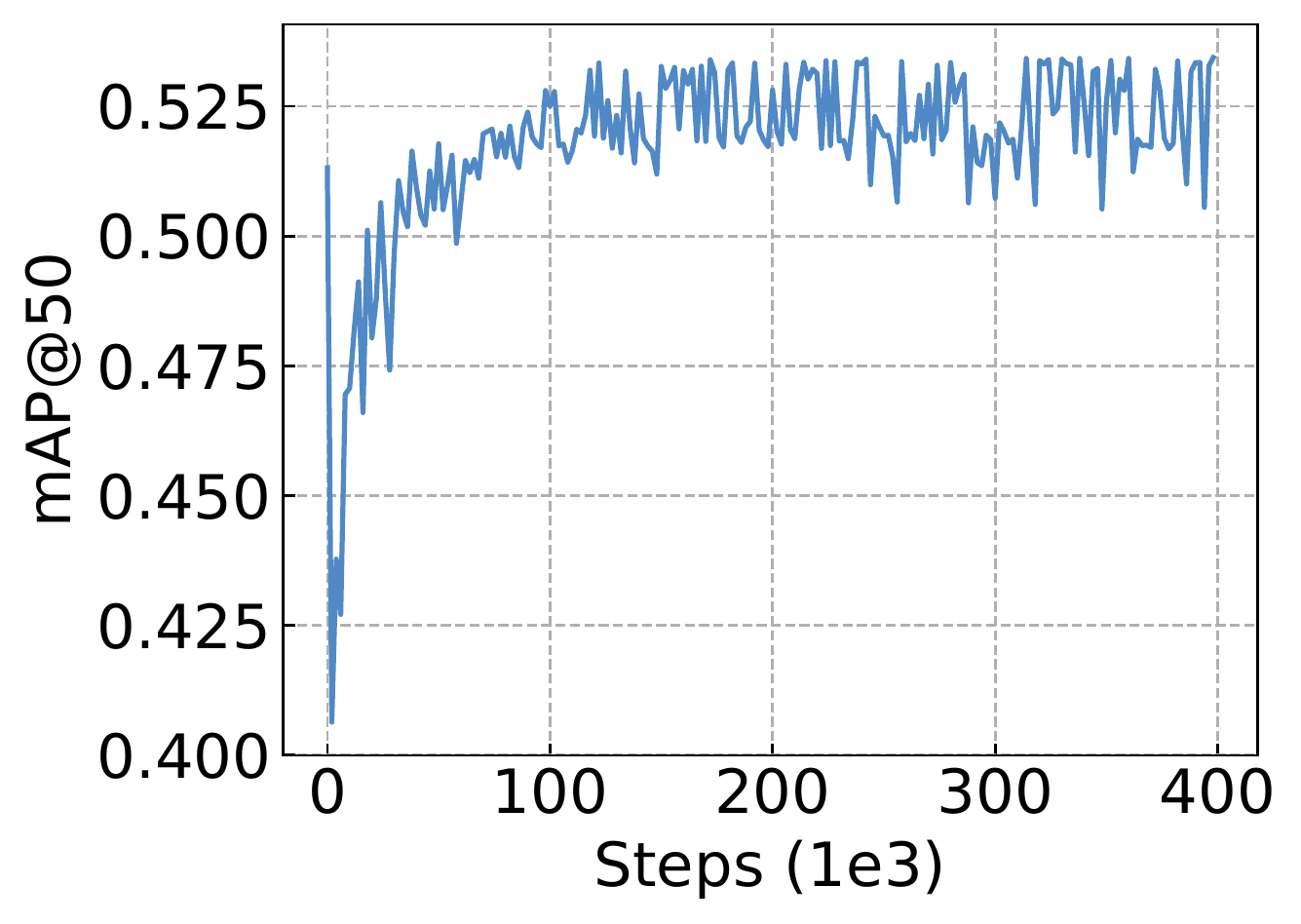}}
\caption{}
\label{map1023}
\end{subfigure}
\hfill
\begin{subfigure}[b]{0.24\textwidth}
\includegraphics[width=\textwidth]{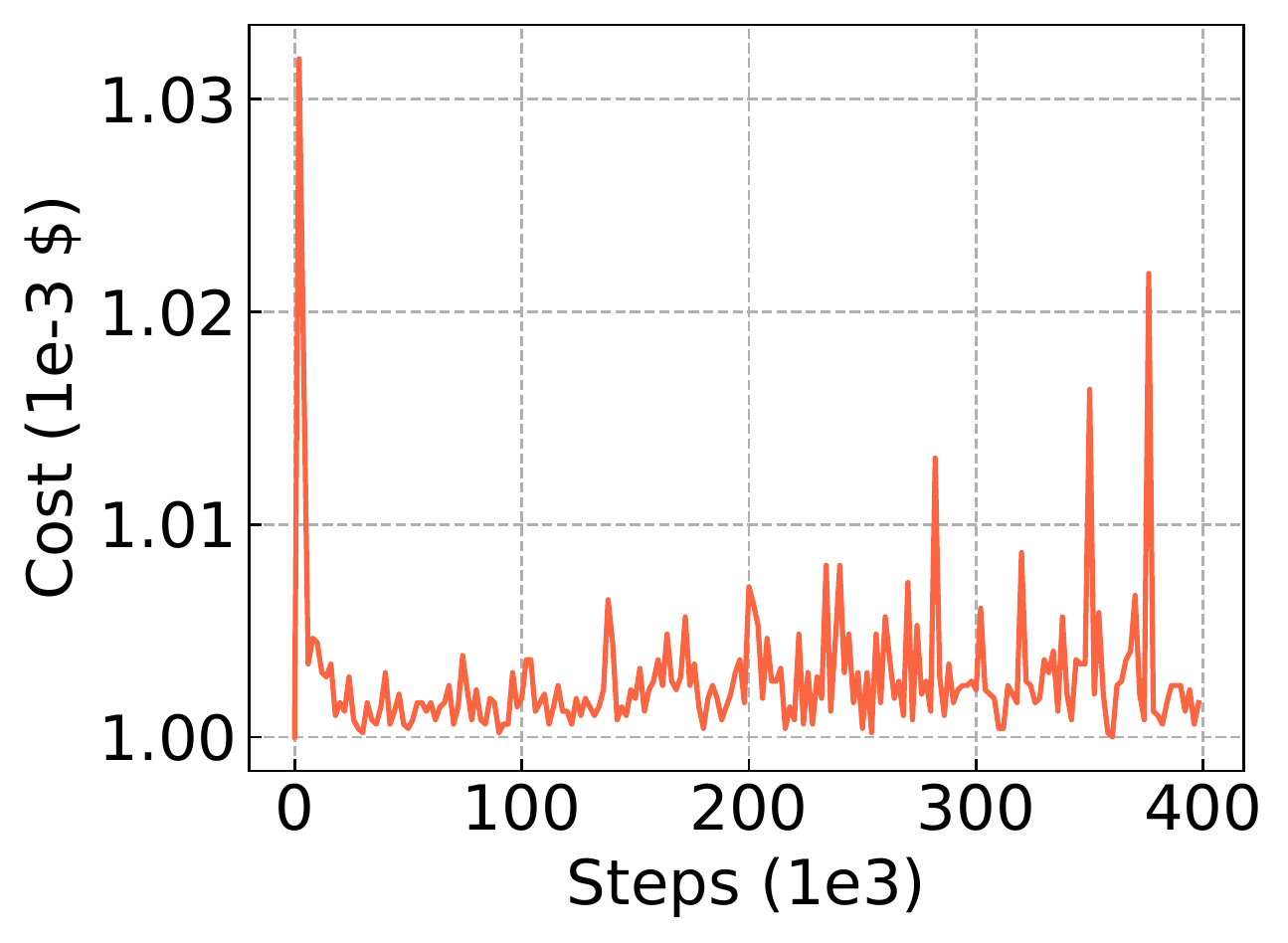}
\caption{}
\label{cost1023}
\end{subfigure}
\caption{Training process with $10$ available MLaaS providers ($1023$ actions). The $mAP@50$ ($AP_{50}$) and cost converges stably.}
\label{fig:scala}
\end{figure}

\textbf{Scalability of combinatorial RL approach.}
To test the scalability of our approach towards a more significant number of MLaaS providers, we add the results of Alibaba Cloud Object Detection \cite{aliyunobjectdet} and synthetic six more MLaaS providers. The details of these simulated MLaaS providers are available in our GitHub repository. 
We index AWS, Azure, Google, Alibaba, and six simulated providers as MLaaS $0$-$9$. In Tab.~\ref{model_pez3}, we find that the ensemble predictions of $10$ MLaaS providers are lower than MLaaS $5$. We suggest that the reason for this phenomenon is because the $AP_{50}$ of MLaaS $5$ is $20\%$-$30\%$ higher compared to the other MLaaS providers, which cannot provide more true positive results and only increases the number of false-positive results in the ensemble predictions. However, as shown in Tab.~\ref{model_pez3}, the $AP_{50}$ of Armol is slightly better than MLaaS $5$ with almost the exact cost, which indicates that although the $AP_{50}$ of the MLaaS providers varies greatly, our algorithm still selects the better combinations. In Fig.~\ref{fig:scala}, our combinatorial RL approach still stably converges with ten providers ($1023$ actions) in both $AP_{50}$ and cost.

\begin{table}[!t]
\caption{Performance metrics of different simulated MLaaS. The unit of cost is $10^{-3}$ USD.}
\begin{center}
\begin{tabular}{|c|c|c||c|c|c|}
\hline
 \textbf{MLaaS} & $\mathbf{AP_{50}}$ & \textbf{Cost} & \textbf{MLaaS} &  $\mathbf{AP_{50}}$ & \textbf{Cost} \\
\hline
 $0$ & $28.88$ & $1.000$ & $5$ & $53.43$ & $1.000$ \\  
\hline
 $1$ & $24.38$ & $1.000$ & $6$ & $20.76$ & $1.000$ \\
\hline
 $2$ & $24.38$ & $1.000$ & $7$ & $51.33$  & $1.000$ \\
\hline
 $3$ & $34.69$ & $1.000$ & $8$ & $25.13$ & $1.000$ \\
\hline
 $4$ & $50.19$  & $1.000$ & $9$ & $34.81$ & $1.000$ \\
\hline
All & $49.29$ & $10.000$ & Armol & $53.44$ & $1.002$ \\
\hline
\end{tabular}
\label{model_pez3}
\end{center}
\end{table}

\section{Related Work} \label{relatedwork}

\subsection{Measurements on Machine Learning Services}
There are previous works that measure the inference accuracy and latency of machine learning models \cite{reddi2020mlperf, zhang2020inferbench}, but these measurements are mainly on user-known models instead of machine learning services. In addition, \cite{yao2017complexity} aims to measure the machine learning training platforms instead of machine learning inference services. There is also a measurement work \cite{liu2018mlbench} on older machine learning services limited to decision trees, SVMs, and multi-layer fully connected neural network services, which is much different from the next generation machine learning services that are now being promoted with models that are transparent to users and in favor of deep learning.

\subsection{Cloud Federation}
Cloud federation comprises services from different providers aggregated in a single pool supporting three basic interoperability features-resource migration, resource redundancy, and complementary resources resp. services \cite{kurze2011cloud}. In the past, cloud federation represents integrating explicit resources, such as storage and compute resources. In contrast, we integrate the implicit resources, which are the training data and model behind the online public machine learning services. Furthermore, the past concept of cloud federation enables further reduction of cost due to partial outsourcing to more cost-efficient regions \cite{giacobbe2015towards}. However, we consider the reduction of cost only after reaching the highest mAP.


\subsection{Reinforcement Learning with Combinatorial Action Spaces}
Discrete, high-dimensional action spaces are common in applications such as natural language processing \cite{DBLP:conf/emnlp/HeOHCGLD16}, text-based applications \cite{DBLP:conf/nips/ZahavyHMMM18} and vehicle routing \cite{DBLP:conf/nips/DelarueAT20}, but they pose a challenge for standard RL algorithms \cite{dulac2019challenges}, because enumerating the action space when choosing the next action from a state becomes impossible. Recent remedies for this problem include selecting the best action from a random sample \cite{DBLP:conf/emnlp/HeOHCGLD16}, approximating the discrete action space with a continuous one \cite{dulac2015deep, DBLP:conf/acl/HeCHGLDO16}, training an additional machine learning model to wean out sub-optimal actions \cite{DBLP:conf/nips/ZahavyHMMM18}, or formulating the action selection problem from each state as a mixed-integer program \cite{DBLP:conf/nips/DelarueAT20}. \cite{zhong2018deep} adds the wolpertinger policy to the edge cache problem, but it is just an application and has no contribution to the policy. Our provider selection approach embeds the continuous action to the nearest neighbor in binary action space. To increase the probability of exploration, we used SAC for training instead of DDPG.

\section{Conclusion} \label{conclusion}
In this paper, we propose Armol, a novel cost-effective MLaaS federation framework that leverages deep combinatorial RL to boost the average precision of federated object detection services so as to minimize the cost. Through our analysis on the predictions of COCO Val $2017$ from AWS Rekognition, Azure Computer Vision, and Google Vision AI, we demonstrate that the mAP of federated MLaaS providers is higher than a single provider, and more MLaaS providers do not mean higher accuracy. Inspired by the recent advances in RL algorithms for combinatorial action spaces, we propose a combinatorial RL-based approach to decide on how to choose the best combination of available MLaaS providers for input. The evaluation further demonstrates the strengths of our approach. 

\section*{Acknowledgment}
Shuzhao Xie thanks Chen Tang, Jiahui Ye, Shiji Zhou, and Wenwu Zhu for their help in making this work possible.
This work is supported in part by NSFC (Grant No. 61872215), and Shenzhen Science and Technology Program (Grant No. RCYX20200714114523079). Yifei Zhu's work is funded by the SJTU Explore-X grant. We would like to thank Tencent for sponsoring the research.

\newpage
\renewcommand*{\bibfont}{\footnotesize}
\printbibliography

@online{aws,
  title = {Amazon Web Services},
  year = {2021},
  url = {https://aws.amazon.com/},
  urldate = {2021-07-18}
}

@online{azure,
  title = {Azure},
  year = {2021},
  url = {https://azure.microsoft.com/},
  urldate = {2021-07-18}
}

@online{google,
  title = {Google Cloud Platform},
  year = {2021},
  url = {https://cloud.google.com/},
  urldate = {2021-07-18}
}

@online{alibaba,
  title = {Alibaba Cloud},
  year = {2021},
  url = {https://www.aliyun.com/},
  urldate = {2021-07-18}
}

@online{awsrekog,
  title = {Amazon Rekognition},
  year = {2021},
  url = {https://aws.amazon.com/rekognition/},
  urldate = {2021-07-18}
}

@online{azurecv,
  title = {Azure Computer Vision},
  year = {2021},
  url = {https://azure.microsoft.com/en-us/services/cognitive-services/computer-vision/},
  urldate = {2021-07-18}
}

@online{googlecv,
  title = {Google Vision AI},
  year = {2021},
  url = {https://cloud.google.com/vision},
  urldate = {2021-07-18}
}

@online{aliyunobjectdet,
  title = {Alibaba cloud object detection},
  year = {2021},
  url = {https://vision.aliyun.com/objectdet},
  urldate = {2021-07-18}
}

@online{mlaasmarket,
  title = {Machine Learning as a Service (MLaaS) Market - Growth, Trends, COVID-19 Impact, and Forecasts (2021 - 2026)},
  year = {2021},
  url = {https://www.reportlinker.com/p06106023/Machine-Learning-as-a-Service-MLaaS-Market-Growth-Trends-COVID-19-Impact-and-Forecasts.html},
  urldate = {2021-07-28}
}

@online{mapvoc2012,
  title = {Average Precision},
  year = {2021},
  url = {http://host.robots.ox.ac.uk/pascal/VOC/voc2012/htmldoc/devkit_doc.html},
  urldate = {2021-07-18}
}

@inproceedings{fujimoto2018addressing,
  title={Addressing function approximation error in actor-critic methods},
  author={Fujimoto, Scott and Hoof, Herke and Meger, David},
  booktitle={ICML 2018},
  pages={1587--1596},
%   year={2018},
%   organization={PMLR}
}

@inproceedings{reddi2020mlperf,
  title={Mlperf inference benchmark},
  author={Reddi, Vijay Janapa and Cheng, Christine and Kanter, David and Mattson, Peter and Schmuelling, Guenther and Wu, Carole-Jean and Anderson, Brian and Breughe, Maximilien and Charlebois, Mark and Chou, William and others},
  booktitle={ACM/IEEE ISCA},
  pages={446--459},
  year={2020}
}

@article{zhang2020inferbench,
  title={InferBench: Understanding Deep Learning Inference Serving with an Automatic Benchmarking System},
  author={Zhang, Huaizheng and Huang, Yizheng and Wen, Yonggang and Yin, Jianxiong and Guan, Kyle},
  journal={arXiv preprint arXiv:2011.02327},
  year={2020}
}

@article{liu2018mlbench,
  title={MLbench: benchmarking machine learning services against human experts},
  author={Liu, Yu and Zhang, Hantian and Zeng, Luyuan and Wu, Wentao and Zhang, Ce},
  journal={Proceedings of the VLDB Endowment},
  volume={11},
  number={10},
  pages={1220--1232},
  year={2018},
  publisher={VLDB Endowment}
}

@inproceedings{lin2014microsoft,
  title={Microsoft coco: Common objects in context},
  author={Lin, Tsung-Yi and Maire, Michael and Belongie, Serge and Hays, James and Perona, Pietro and Ramanan, Deva and Doll{\'a}r, Piotr and Zitnick, C Lawrence},
  booktitle={ECCV 2014},
  pages={740--755},
  %year={2014},
  organization={Springer}
}

@inproceedings{yao2017complexity,
  title={Complexity vs. performance: empirical analysis of machine learning as a service},
  author={Yao, Yuanshun and Xiao, Zhujun and Wang, Bolun and Viswanath, Bimal and Zheng, Haitao and Zhao, Ben Y},
  booktitle={Proc. IMC 2017},
  pages={384--397},
  %year={2017}
}

@article{kurze2011cloud,
  title={Cloud federation},
  author={Kurze, Tobias and Klems, Markus and Bermbach, David and Lenk, Alexander and Tai, Stefan and Kunze, Marcel},
  journal={Cloud Computing},
  volume={2011},
  pages={32--38},
  year={2011},
  publisher={Citeseer}
}

@article{giacobbe2015towards,
  title={Towards energy management in cloud federation: a survey in the perspective of future sustainable and cost-saving strategies},
  author={Giacobbe, Maurizio and Celesti, Antonio and Fazio, Maria and Villari, Massimo and Puliafito, Antonio},
  journal={Computer Networks},
  volume={91},
  pages={438--452},
  year={2015},
  publisher={Elsevier}
}

@article{10.1145/219717.219748,
author = {Miller, George A.},
title = {WordNet: A Lexical Database for English},
year = {1995},
issue_date = {Nov. 1995},
publisher = {Association for Computing Machinery},
address = {New York, NY, USA},
volume = {38},
number = {11},
journal = {Commun. ACM},
month = nov,
pages = {39–41},
numpages = {3}
}

@inproceedings{hosang2017learning,
  title={Learning non-maximum suppression},
  author={Hosang, Jan and Benenson, Rodrigo and Schiele, Bernt},
  booktitle={IEEE CVPR 2017},
  pages={4507--4515},
  %year={2017}
}

@inproceedings{bodla2017soft,
  title={Soft-NMS--improving object detection with one line of code},
  author={Bodla, Navaneeth and Singh, Bharat and Chellappa, Rama and Davis, Larry S},
  booktitle={ICCV 2017},
  pages={5561--5569},
  %year={2017}
}

@article{solovyev2019weighted,
  title={Weighted boxes fusion: ensembling boxes for object detection models},
  author={Solovyev, Roman and Wang, Weimin and Gabruseva, Tatiana},
  journal={arXiv preprint arXiv:1910.13302},
  year={2019}
}

@inproceedings{DBLP:conf/ecai/Casado-GarciaH20,
  author    = {Angela Casado{-}Garcia and
               Jonathan Heras},
  title     = {Ensemble Methods for Object Detection},
  booktitle = {{ECAI} 2020},
  pages     = {2688--2695},
  %year      = {2020}
}

@article{dulac2015deep,
  title={Deep reinforcement learning in large discrete action spaces},
  author={Dulac-Arnold, Gabriel and Evans, Richard and van Hasselt, Hado and Sunehag, Peter and Lillicrap, Timothy and Hunt, Jonathan and Mann, Timothy and Weber, Theophane and Degris, Thomas and Coppin, Ben},
  journal={arXiv preprint arXiv:1512.07679},
  year={2015}
}

@inproceedings{DBLP:conf/acl/HeCHGLDO16,
  author    = {Ji He and
               Jianshu Chen and
               Xiaodong He and
               Jianfeng Gao and
               Lihong Li and
               Li Deng and
               Mari Ostendorf},
  title     = {Deep Reinforcement Learning with a Natural Language Action Space},
  booktitle = {Proc. ACL 2016},
  publisher = {The Association for Computer Linguistics},
  %year      = {2016}
}

@inproceedings{DBLP:conf/emnlp/HeOHCGLD16,
  author    = {Ji He and
               Mari Ostendorf and
               Xiaodong He and
               Jianshu Chen and
               Jianfeng Gao and
               Lihong Li and
               Li Deng},
  title     = {Deep Reinforcement Learning with a Combinatorial Action Space for
               Predicting Popular Reddit Threads},
  booktitle = {Proc. EMNLP 2016},
  pages     = {1838--1848},
  %year      = {2016}
}

@inproceedings{DBLP:conf/nips/ZahavyHMMM18,
  author    = {Tom Zahavy and
               Matan Haroush and
               Nadav Merlis and
               Daniel J. Mankowitz and
               Shie Mannor},
  title     = {Learn What Not to Learn: Action Elimination with Deep Reinforcement
               Learning},
  booktitle = {NeurIPS 2018},
  pages     = {3566--3577},
  %year      = {2018}
}

@inproceedings{DBLP:conf/nips/DelarueAT20,
  author    = {Arthur Delarue and
               Ross Anderson and
               Christian Tjandraatmadja},
  title     = {Reinforcement Learning with Combinatorial Actions: An Application
               to Vehicle Routing},
  booktitle = { NeurIPS 2020},
  %year      = {2020}
}

@inproceedings{zhong2018deep,
  title={A deep reinforcement learning-based framework for content caching},
  author={Zhong, Chen and Gursoy, M Cenk and Velipasalar, Senem},
  booktitle={IEEE CISS 2018},
  pages={1--6},
  %year={2018},
  %organization={IEEE}
}

@article{dulac2019challenges,
  title={Challenges of real-world reinforcement learning},
  author={Dulac-Arnold, Gabriel and Mankowitz, Daniel and Hester, Todd},
  journal={arXiv preprint arXiv:1904.12901},
  year={2019}
}

@article{schulman2017proximal,
  title={Proximal policy optimization algorithms},
  author={Schulman, John and Wolski, Filip and Dhariwal, Prafulla and Radford, Alec and Klimov, Oleg},
  journal={arXiv preprint arXiv:1707.06347},
  year={2017}
}

@inproceedings{haarnoja2018soft,
  title={Soft actor-critic: Off-policy maximum entropy deep reinforcement learning with a stochastic actor},
  author={Haarnoja, Tuomas and Zhou, Aurick and Abbeel, Pieter and Levine, Sergey},
  booktitle={ICML 2018},
  pages={1861--1870},
%   year={2018},
%   organization={PMLR}
}

@article{SpinningUp2018,
    author = {Achiam, Joshua},
    title = {{Spinning Up in Deep Reinforcement Learning}},
    year = {2018}
}
\end{document}